\documentclass[conference]{IEEEtran}
\IEEEoverridecommandlockouts

\usepackage{cite}
\usepackage{amsmath,amssymb,amsfonts}
\usepackage{algorithmic}
\usepackage{graphicx}
\usepackage{textcomp}
\usepackage{xcolor}
\usepackage{amsfonts} 
\usepackage{amsthm}
\usepackage{amsmath}
\usepackage{algorithm}
\usepackage{algorithmic}
\usepackage{subcaption}
\usepackage[left=1.62cm,right=1.62cm,top=1.85cm]{geometry}
\usepackage{multirow}
\usepackage{booktabs}
\usepackage{url}
\def\BibTeX{{\rm B\kern-.05em{\sc i\kern-.025em b}\kern-.08em
    T\kern-.1667em\lower.7ex\hbox{E}\kern-.125emX}}
\begin{document}

\title{{Power to the Clients: Federated Learning in a Dictatorship Setting}
}

\author{
\IEEEauthorblockN{Mohammadsajad Alipour, Mohammad Mohammadi Amiri}
\IEEEauthorblockA{
\textit{Department of Computer Science},
\textit{Rensselaer Polytechnic Institute},
Troy, NY 12180, USA \\
\{alipom, mamiri\}@rpi.edu
}\thanks{The online preprint of this paper, including the appendix, is available at arxiv.org/abs/2510.22149}
}

\maketitle

\begin{abstract}
Federated learning (FL) has emerged as a promising paradigm for decentralized model training, enabling multiple clients to collaboratively learn a shared model without exchanging their local data. However, the decentralized nature of FL also introduces vulnerabilities, as malicious clients can compromise or manipulate the training process. In this work, we introduce \textbf{dictator clients}, a novel, well-defined, and analytically tractable class of malicious participants capable of entirely erasing the contributions of all other clients from the server model, while preserving their own. We propose concrete attack strategies that empower such clients and systematically analyze their effects on the learning process. Furthermore, we explore complex scenarios involving multiple dictator clients, including cases where they collaborate, act independently, or form an alliance in order to ultimately betray one another. For each of these settings, we provide a theoretical analysis of their impact on the global model's convergence. Our theoretical algorithms and findings about the complex scenarios including multiple dictator clients are further supported by empirical evaluations on both computer vision and natural language processing benchmarks.
\end{abstract}


\section{Introduction}
Federated learning (FL) \cite{mcmahan2017communication} is a distributed learning paradigm in which model training is performed collaboratively by a set of clients. In centralized FL, a global server broadcasts the current model to all clients, each of which updates the model using its local dataset and sends back the resulting gradients to the server. The server then aggregates these gradients to update the global model. This approach accelerates training by distributing computation across multiple machines, while also enhancing data privacy since clients share only gradients, not their raw data. FL is especially well-suited for privacy-sensitive applications, such as training on confidential medical records across hospitals.

Despite its advantages, FL remains vulnerable to malicious behavior by the participating clients. \textit{Byzantine clients} are adversarial participants that disrupt the training process by sending arbitrary or manipulated updates to the central server \cite{lamport2019byzantine}\cite{NIPS2017_f4b9ec30blanch}. The presence of such adversaries can significantly degrade model performance, making Byzantine robustness a critical area of study \cite{li2019rsa,wu2020federated,shejwalkar2021manipulating,guerraoui2018hidden,xie2018generalized,xie2022game}. Moreover, several studies have demonstrated the possibility of backdoor attacks in FL via collusion attacks where multiple malicious clients coordinate their actions to inject hidden triggers into the global model in FL \cite{liu2024act,ranjan2022securing,xiao2022sca,pmlr-v108-bagdasaryan20a}. These clients may exchange information and strategically craft updates that steer the aggregated model toward a compromised state. 

However, the majority of existing literature primarily focuses on defending against Byzantine clients, while comparatively little attention has been given to characterizing specific and well-defined behaviors of Byzantine clients that have a different specific goal\textemdash especially in exploring diverse scenarios involving their presence within the system. In FL, a malicious client may aim to impose the statistical properties or specific patterns of its own dataset onto the global model. Such a client effectively attempts to \textit{dictate} the final model by aligning it more closely with its local data distribution. This behavior may serve various objectives, such as improving performance on a target task, biasing global model's decisions toward a desired objective, embedding backdoors, or degrading the model’s generalization on other clients' data. By exploiting vulnerabilities in the model aggregation process, especially when contributions are blindly averaged or insufficiently audited, a malicious client can steer the training dynamics to serve its own objectives, ultimately dominating the global model's behavior.

In this work, we introduce a novel and formally defined class of Byzantine clients in FL, characterized by precise assumptions about their knowledge of the system and limitations. In contrast to prior studies, which often assumed omniscient or overly powerful adversaries, we consider malicious clients with only minimal communication capabilities among themselves. These clients lack visibility into the internal structure of the global model and have no information about the data or updates of benign clients. By clearly bounding their capabilities, our framework offers a more realistic and fine-grained understanding of adversarial behavior in practical FL environments. 

The goal of these malicious clients is to preserve their own influence on the final global model while entirely eliminating the contributions of all other participants\textemdash as if the benign clients had never been involved in the training process. We refer to such independent malicious clients as \textit{dictator} clients due to their unilateral domination of the model. When multiple such clients coordinate via their limited communication link to jointly dominate training, we refer to them as \textit{collaborative dictator} clients. We show that these clients do not require any privileged access to the server or any external metadata\textemdash making their attack strategies particularly concerning from a security perspective.

To demonstrate the feasibility of this threat, we develop a series of algorithms that enable malicious clients to achieve their goals within the defined constraints. Our theoretical findings are further supported by empirical results, which validate the effectiveness of the proposed attack strategies. Beyond isolated attacks, we also investigate complex and previously underexamined dynamics that arise among malicious clients themselves. For example, we examine scenarios in which all participants in the system act as dictators, as well as cases where collaborative dictator clients can betray one another within their own partnership. These scenarios reveal internal conflicts among adversaries and broaden the understanding of multi-agent adversarial behavior in FL. The practical implications of dictator clients are also discussed in more detail in Appendix \ref{implics}.

\section{Related Work}
The distributed nature of FL, combined with the server's limited visibility into local training processes, makes it vulnerable to various security threats posed by malicious or compromised clients \cite{10.1145/3543873.3587681trust}. In this section, we review existing literature across three major category of attacks\textemdash Byzantine attacks, backdoor attacks, and collusion attacks.

\subsection*{Byzantine Attacks}
Byzantine attacks pose a fundamental threat in distributed systems including FL, where a subset of clients, known as \textit{Byzantine clients}, arbitrarily deviate from the prescribed protocol by submitting malicious or anomalous updates to the central server \cite{lamport2019byzantine}. The goals of such attacks typically include degrading the global model's performance or preventing convergence \cite{NIPS2017_f4b9ec30blanch}.
Attack strategies vary in complexity, ranging from simple approaches such as random noise injection or submitting zero gradients to more sophisticated methods like sign-flipping \cite{10.1007/978-981-99-7032-2_3sybilmitig,Shen_Huang_Wan_Ye_2025}. Advanced attacks are often crafted to evade specific defenses, making them challenging to detect and mitigate \cite{shejwalkar2021manipulating,baruch2019little}.

\subsection*{Backdoor Attacks}
Backdoor attacks (also known as Trojan attacks) are a more insidious threat in FL where attackers aim to embed hidden malicious behavior into the global model \cite{gu2017badnets,li2022backdoor}. An attacker, typically controlling one or more clients, manipulates their local dataset or model updates to create a "backdoor trigger"\textemdash a specific pattern or feature (e.g., a small patch in an image, a specific phrase in text). The compromised global model performs normally on clean inputs but exhibits attacker-chosen behavior, such as misclassification, when the trigger is present. These attacks can be implemented through various strategies, including \textbf{data poisoning}, where labels are manipulated for samples containing the trigger, and \textbf{model poisoning}, where malicious updates are directly crafted to influence model behavior \cite{pmlr-v108-bagdasaryan20a,Xie2020DBA:}. Triggers may be static and predefined \cite{pmlr-v108-bagdasaryan20a} or dynamically generated using optimization techniques to make them more subtle and difficult to detect \cite{NEURIPS2023_c07d71ffadap}. Comprehensive surveys on backdoor attacks and defenses in FL can be found in \cite{NGUYEN2024107166}.

\subsection*{Collusion Attacks}
Collusion attacks occur when multiple malicious clients coordinate their actions to enhance the effectiveness of the attacks or bypass defenses designed for independent attackers. Colluding attackers can amplify the impact of Byzantine or backdoor attacks. For example, multiple Byzantine clients might coordinate their updates to overwhelm Byzantine-resilient aggregation rules that assume the number of attackers are limited \cite{Xie2020DBA:}. Similarly, colluding clients can implement distributed backdoor attacks, where each attacker contributes a part of the malicious update, making individual contributions appear benign while collectively embedding a backdoor into the global model \cite{Lyu_Han_Wang_Liu_Wang_Liu_Zhang_2023}. More advanced and specific collusion strategies include \textit{alternating attacks} and \textit{stealthy collusion}. In alternating (on-off) attacks, malicious clients alternate between benign and malicious behavior to build reputation or evade history-based detection \cite{JMLR:v24:22-0014}. In stealthy collusion attacks, attackers coordinate to make their cumulative malicious impact significant while keeping individual updates close to benign ones to evade detection \cite{10638807sparse}. Such attacks aim for sparsity and stealthiness.

While prior research has primarily focused on degrading model utility or embedding backdoors, our work introduces and formalizes a new adversarial paradigm: \textit{dictator clients}\textemdash malicious participants whose goal is not to harm performance but to fully preserve their own contribution to the global model while completely erasing the influence of other clients. Unlike traditional Byzantine or backdoor attacks, dictator clients aim to \textit{bias} the learning outcome toward their local objectives without necessarily compromising overall model accuracy. Moreover, we investigate nuanced interaction dynamics among multiple dictator clients, including collaboration, conflict, and strategic deception. To the best of our knowledge, this is the first systematic exploration of such influence-preserving and interaction-aware attacks, revealing a novel and underexplored threat model in FL.

\section{Problem Formulation and Preliminaries}
We consider a centralized FL setting in which, during each communication round, a central server broadcasts the current model weights to all clients. Each client then performs stochastic gradient descent on the loss function on its local dataset to compute an update. These local updates are sent back to the server, which aggregates them\textemdash most commonly through simple averaging\textemdash and applies a global gradient descent step scaled by a predefined learning rate. To enable a more precise formulation and analysis of the attacks, we assume that the server aggregates updates from all clients in every round\textemdash an assumption that commonly holds in cross-silo FL settings \cite{10373828crosssilo}. We defer to future work the exploration of FL variants that either allow partial client participation or permit clients to perform several local updates before aggregation.

Let $\theta_t$ denote the global model weights maintained by the server at iteration $t$, and let $\mathcal{N} = \{1, 2, \dots, N\}$ represent the set of $N$ participating clients. For each client $n \in \mathcal{N}$, let $\nabla \mathcal{L}_n(\theta_t)$ denote the gradient of its local loss function with respect to the current model $\theta_t$. The server updates the global model at each round after collecting the gradients from all clients as follows:
\begin{align}\label{prelim-normal}
    {\theta_{t+1}}=\theta_t - \eta \sum\nolimits_{n=1}^N{\nabla{\mathcal{L}_n(\theta_t)}},
\end{align}
where $\eta > 0$ denotes the server-side learning rate. The global model is initialized as $\theta_0$ at the server and distributed to all clients at the beginning of training.

We further define a hypothetical baseline scenario where only a single client $m \in \mathcal{N}$ participates in the learning process. Let $\hat{\theta}^m_t$ denote the model weights at iteration $t$ in this single-client scenario. The corresponding update rule simplifies to $
    \hat{\theta}^m_{t+1}=\hat{\theta}^m_{t} - \eta {\nabla{\mathcal{L}_m(\hat{\theta}^m_{t})}},
$
with initialization $\hat{\theta}^m_0 = \theta_0$.
We further generalize this formulation to a subset of clients. Let $\mathcal{P} \subset \mathcal{N}$ denote a subset of $P$ clients, where $1 < P < N$. We define $\hat{\theta}^{\mathcal{P}}_t$ as the model weights at iteration $t$ when only clients in $\mathcal{P}$ participate in training. The update rule for this partial participation scenario is given by $
    \hat{\theta}^\mathcal{P}_{t+1}=\hat{\theta}^\mathcal{P}_{t} - \eta \sum\nolimits_{k\in \mathcal{P}}{\nabla{\mathcal{L}_k(\hat{\theta}^\mathcal{P}_{t})}},
$
with initialization $\hat{\theta}^{\mathcal{P}}_0 = \theta_0$. Next, we introduce scenarios involving dictator clients in FL, including both single-dictator and multi-dictator cases. We describe how these clients modify their local updates to achieve their objectives. Specifically, a single dictator client aims to steer the global model's updates and convergence to follow $\hat{\theta}^m_{t+1}$, while a group of coordinated dictator clients (collaborative dictators) seeks to enforce convergence toward $\hat{\theta}^\mathcal{P}_{t+1}$, effectively overriding the standard FL update rule in Eq.~\eqref{prelim-normal}.

\section{Dictator Client Scenarios}
In this section, we propose algorithms that enable clients to become dictators\textemdash retaining their own contributions to the global model while eliminating those of others. We begin with the case of a single dictator client in Section \ref{one} and then extend to scenarios involving multiple collaborating dictator clients in Section \ref{partner}. Figure~\ref{fig:overview} (in Appendix~\ref{visu}) illustrates different dictator client scenarios compared with standard FL.

\subsection{Single Dictator Client}\label{one}
In this section, we demonstrate how a single dictator client can craft its updates to entirely nullify the contributions of all other clients while preserving its own influence on the global model. We assume that the dictator client knows only the server's learning rate and requires no additional information. Notably, as shown in Appendix~\ref{learningrate}, even this assumption can be relaxed, as the learning rate can be numerically estimated after a single iteration. Suppose client $m \in \mathcal{N}$ such that $1\leq m\leq N$ is the designated dictator client and only knows server's learning rate $\eta$. At iteration $0$, the server broadcasts the initial model $\theta_0$ to all clients, which each use to compute their local gradients. Upon receiving these gradients, the server updates the global model as $\theta_1=\theta_0-\eta \sum\nolimits_{n=1}^N \nabla{\mathcal{L}_n(\theta_0)}$.
In the next iteration, the server broadcasts $\theta_1$ to all clients. Each client except client $m$, computes and sends their gradient with respect to $\theta_1$. Meanwhile, client $m$ retains a local copy of the initial server model $\theta_0$ from the previous iteration. Using this, it computes a hypothetical model update, denoted by $\hat{\theta}^m_1$, which represents the model that would have resulted if only client $m$'s gradient had been used in the first iteration. This is computed as:
\begin{align}\label{theta1mal}
    \hat{\theta}^m_1=\theta_0-\eta \nabla{\mathcal{L}_m(\theta_0)}.
\end{align}
The dictator client $m$ sends a carefully crafted update $M_1$ instead of its actual gradient $\nabla{\mathcal{L}_m(\theta_1)}$ to delete the contribution of all other clients from the previous iteration and preserve only its own contribution. This manipulated update is defined as $    M_1=\nabla{\mathcal{L}_m(\hat{\theta}^m_1) - \left(\frac{\theta_0-\theta_1}{\eta}-\nabla\mathcal{L}_m(\theta_0)\right)}.
$
Here, the term $\frac{\theta_0-\theta_1}{\eta}$ reconstructs the aggregate gradient used by the server in the first round, allowing client $m$ to effectively cancel out the influence of all other clients while steering the update toward its own objective.
We now analyze the updated global model $\theta_2$ after the server aggregates all client updates in the second iteration:
\begin{align*}
    \theta_2&=\theta_1-\eta\left(M_1+{\sum\nolimits_{n=1,n\neq{m}}^N\nabla{\mathcal{L}_n(\theta_1)}}\right)
    \\
    &=\theta_1-\eta(\nabla{\mathcal{L}_m(\hat{\theta}^m_1)}-(\frac{\theta_0-\theta_1}{\eta}-\nabla{\mathcal{L}_m(\theta_0))}\\&\quad+{\sum\nolimits_{n=1,n\neq{m}}^N\nabla{\mathcal{L}_n(\theta_1)}})
    \\
    &=\theta_0-\eta \nabla{\mathcal{L}_m(\theta_0)} - \eta(\nabla{\mathcal{L}_m(\hat{\theta}^m_1)}\\&\quad+{\sum\nolimits_{n=1,n\neq{m}}^N\nabla{\mathcal{L}_n(\theta_1)}}).
\end{align*}
Now, substituting from Eq.~\eqref{theta1mal}, we can express $\theta_2$ as $
    \theta_2=\hat{\theta}^m_1-\eta\left(\nabla{\mathcal{L}_m(\hat{\theta}^m_1)}+{\sum\nolimits_{n=1,n\neq{m}}^N\nabla{\mathcal{L}_n(\theta_1)}}\right).
$
This demonstrates that by sending the carefully crafted update $M_1$, client $m$ effectively nullifies the contributions of all other clients from the previous iteration while retaining its own gradient contribution. In doing so, the server's model state is steered toward the single-client trajectory $\hat{\theta}^m_1$ instead of the standard FL update. To generalize this strategy for any round $t$, client $m$ maintains a local model $\hat{\theta}^m_t$, which is updated independently as $    \hat{\theta}^m_t=\hat{\theta}^m_{t-1}-\eta\nabla{\mathcal{L}_m(\hat{\theta}^m_{t-1})}$.
We define $M_t$ as the update that the dictator client $m$ sends to the server at iteration $t$ as:   
\begin{align*}
    M_t=\nabla{\mathcal{L}_m(\hat{\theta}^m_t)}-\left(\frac{\hat{\theta}^m_{t-1}-\theta_t}{\eta}-\nabla{\mathcal{L}_m(\hat{\theta}^m_{t-1}})\right).
\end{align*}
Now, we analyze the server's model update at iteration $t+1$ after aggregating all client updates:
\begin{align*}
    \theta_{t+1}&=\theta_t-\eta\left(M_t+{\sum\nolimits_{n=1,n\neq{m}}^N\nabla{\mathcal{L}_n(\theta_t)}}\right)
    \\
    &=\theta_t-\eta(\nabla{\mathcal{L}_m(\hat{\theta}^m_t)}-(\frac{\hat{\theta}^m_{t-1}-\theta_t}{\eta}-\nabla{\mathcal{L}_m(\hat{\theta}^m_{t-1}))}\\&\quad+{\sum\nolimits_{n=1,n\neq{m}}^N\nabla{\mathcal{L}_n(\theta_t)}})
    \\
    &=\hat{\theta}^m_{t-1}-\eta \nabla{\mathcal{L}_m(\hat{\theta}^m_{t-1})} - \eta(\nabla{\mathcal{L}_m(\hat{\theta}^m_t)}\\&\quad+{\sum\nolimits_{n=1,n\neq{m}}^N\nabla{\mathcal{L}_n(\theta_t)}}).
\end{align*}
Substituting $\hat{\theta}^m_t$, it follows that $\theta_{t+1}=\hat{\theta}^m_t-\eta\left(\nabla{\mathcal{L}_m(\hat{\theta}^m_t)}+{\sum\nolimits_{n=1,n\neq{m}}^N\nabla{\mathcal{L}_n(\theta_t)}}\right)$.
After $T$ rounds of training, the final model weights $\theta^*$ will be:
\begin{align}
    \theta^*&=\hat{\theta}^m_T-\eta(\nabla{\mathcal{L}_m(\hat{\theta}^m_T)}+{\sum\nolimits_{n=1,n\neq{m}}^N\nabla{\mathcal{L}_n(\theta_T)}})\\&=\hat{\theta}^m_T-\eta\nabla{\mathcal{L}_m(\hat{\theta}^m_T)}-\eta{\sum\nolimits_{n=1,n\neq{m}}^N\nabla{\mathcal{L}_n(\theta_T)}}\notag\\
    &=\hat{\theta}^m_{T+1}-\eta{\sum\nolimits_{n=1,n\neq{m}}^N\nabla{\mathcal{L}_n(\theta_T)}} \approx \hat{\theta}^m_{T+1}\label{fin}.
\end{align}
This final expression shows that the dictator client drives the model toward its own trajectory $\hat{\theta}^m_{T+1}$, effectively overriding the influence of other clients up to a residual term.
As shown in Eq.~\eqref{fin}, the exact final weights under our method are given by $ \theta^*=\hat{\theta}^m_{T+1}-\eta{\sum\nolimits_{n=1,n\neq{m}}^N\nabla{\mathcal{L}_n(\theta_T)}}$,
where $\hat{\theta}^m_{T+1}$ represents the weights for the final iteration if only the dictator client $m$ had participated in training, as if no other client existed. The residual term $\eta{\sum\nolimits_{n=1,n\neq{m}}^N\nabla{\mathcal{L}_n(\theta_T)}}$ captures the contributions from the other clients in the final iteration. In practice, this term is negligible, as it stems from a single round of updates and has minimal impact on the final model\textemdash especially when the model produced by the dictator client is robust to such perturbations. Our empirical results in Section \ref{exp} further support the insignificance of this residual term on the dictator client's objective. Algorithm \ref{One-malicious-client} (in Appendix~\ref{appalg1}) outlines the complete procedure for a client to act as a dictator.

\begin{table*}
    \scriptsize
    \centering
    \begin{tabular}{c|ccccc||ccccc}
        \toprule
        \textbf{Method} 
        & \multicolumn{5}{c||}{\textbf{MNIST}} 
        & \multicolumn{5}{c}{\textbf{CIFAR-10}} \\
        \cmidrule{2-11}
        & [\textbf{0,1}] & [\textbf{2,3}] & [\textbf{4,5}] & [\textbf{6,7}] & [\textbf{8,9}]
        & [\textbf{0,1}] & [\textbf{2,3}] & [\textbf{4,5}] & [\textbf{6,7}] & [\textbf{8,9}] \\
        \midrule
        Regular FL 
        & 96.18 & 79.25 & 66.84 & 88.12 & 66.38
        & 39.04 & 12.51 & 31.07 & 23.74 & 52.59 \\
        \midrule
        Dictator client: $1$ 
        & 99.63 & 0.00 & 0.00 & 0.00 & 0.00
        & 73.65 & 0.00 & 0.00 & 0.00 & 0.00 \\
        Dictator client: $2$ 
        & 0.00 & 93.92 & 0.00 & 0.00 & 0.00
        & 0.00 & 65.19 & 0.00 & 0.00 & 0.00 \\
        Dictator client: $3$ 
        & 0.00 & 0.00 & 97.43 & 0.00 & 0.00
        & 0.00 & 0.00 & 66.51 & 0.00 & 0.00 \\
        Dictator client: $4$ 
        & 0.00 & 0.00 & 0.00 & 98.91 & 0.00
        & 0.00 & 0.00 & 0.00 & 73.98 & 0.00 \\
        Dictator client: $5$ 
        & 0.00 & 0.00 & 0.00 & 0.00 & 94.42
        & 0.00 & 0.00 & 0.00 & 0.00 & 77.06 \\
        \midrule
        Dictator clients: $2$,$3$
        & 0.00 & 88.19 & 87.80 & 0.00 & 0.00
        & 0.00 & 35.08 & 43.17 & 0.00 & 0.00 \\
        Dictator clients: $2$,$3$,$4$
        & 0.00 & 84.87 & 80.22 & 94.19 & 0.00
        & 0.00 & 18.38 & 40.02 & 46.05 & 0.00 \\
        \bottomrule
    \end{tabular}
    \caption{Performance of the global model on each local dataset for MNIST and CIFAR-10 datasets and the single dictator client and collaborative dictator clients scenarios.}
    \label{tab:resultsOne}
\end{table*}

\subsection{Collaborative Dictator Clients}\label{partner}
In this section, we extend the single dictator client scenario to a group of $P$ dictator clients acting in coordination. As illustrated in Figure~\ref{fig:overview}(c) (in Appendix~\ref{visu}), these clients collaborate to suppress the influence of all others while preserving their own contributions, relying only on inter-client communication. As discussed in Appendix~\ref{learningrate}, they do not require prior knowledge of the server's learning rate\textemdash it can be accurately estimated after a single training round. Let $\mathcal{P} \subset \mathcal{N}$ denote the set of $P$ collaborating dictator clients, where $1 < P < N$, capable of communicating with each other. These clients coordinate their updates according to Algorithm~\ref{Partner-malicious-client} (in Appendix~\ref{appalg2}) so that the global model evolves as if only they had participated in training. 
Each client in $\mathcal{P}$ maintains a synchronized local model, denoted as $\hat{\theta}^\mathcal{P}_t$, representing the model state at iteration $t$ under their exclusive contributions from the start. At each round, every dictator client $k\in\mathcal{P}$ submits a crafted update to the server, effectively nullifying the impact of the remaining $N - P$ clients in $\mathcal{N} \setminus \mathcal{P}$, defined as $
M^k_t=\nabla{\mathcal{L}_k(\hat{\theta}^\mathcal{P}_t)}-\left( \frac{\hat{\theta}^\mathcal{P}_{t-1}-\theta_t}{P\eta}-\nabla{\mathcal{L}_k(\hat{\theta}^\mathcal{P}_{t-1}})\right),\forall k\in \mathcal{P}$.
Afterwards, the clients exchange gradients to jointly compute the next local model state, $\hat{\theta}^\mathcal{P}_{t+1}$. As long as all $P$ clients in $\mathcal{P}$ follow this protocol and continues to share gradients for updating the collective local model, the global model will converge as if only the clients in $\mathcal{P}$ had trained it. A formal proof of this outcome is provided in Appendix~\ref{proof-partner}. Next, we turn our attention to more intricate interactions that emerge in FL systems with the presence of such dictator clients.

\section{Experiments}\label{exp}
In this section, we empirically evaluate the effectiveness of our proposed attack algorithms across both computer vision and natural language processing (NLP) tasks. For our main experiments, we focus on image classification using the MNIST \cite{lecun1998gradient} and CIFAR10 \cite{krizhevsky2009learning} datasets with a simple convolutional neural network (CNN) as the global model. To maintain consistency with our theoretical framework, all experiments are conducted using stochastic gradient descent (SGD) as the optimizer. We simulate a FL environment with five clients, each assigned a disjoint and non-overlapping subset of the training data to create a highly not independent and identically distributed (non-IID) setting. Specifically, the training set is partitioned such that client 1 receives samples with labels 0 and 1, client 2 with labels 2 and 3, and so on, ensuring that each client maintains only two unique classes. Additional results for NLP tasks are provided in Appendix~\ref{nlp}. 

\subsection{Single Dictator Client}

\begin{figure}
    \centering
    \includegraphics[width=0.42\linewidth]{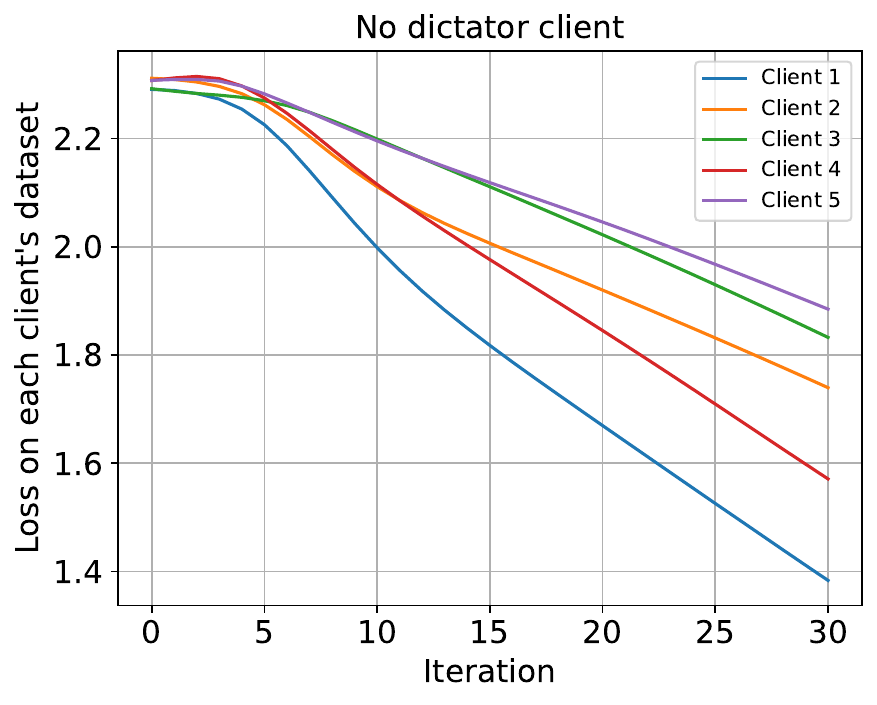}
    \includegraphics[width=0.42\linewidth]{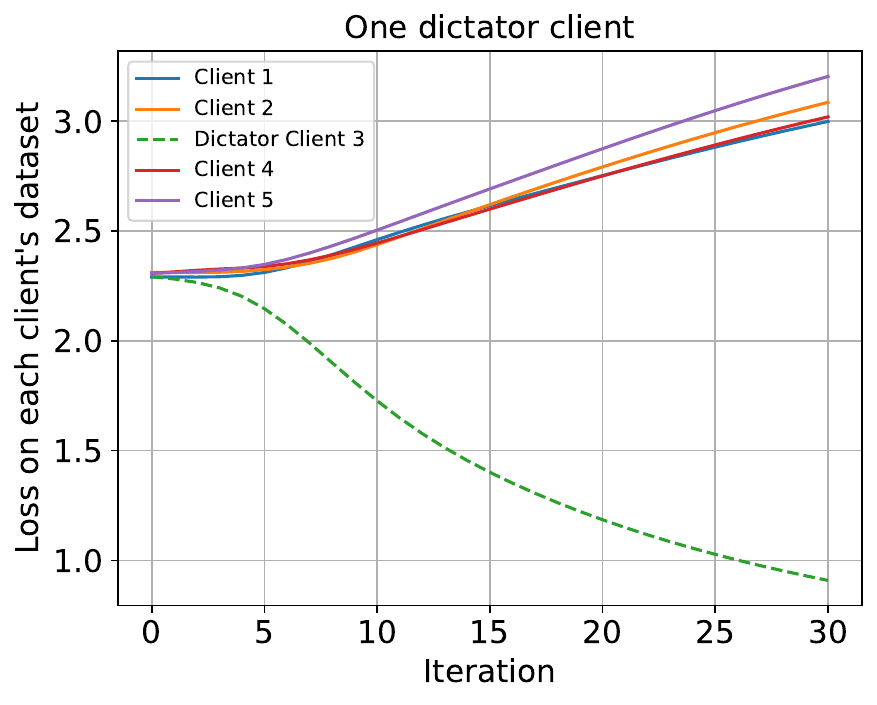}
    \caption{Loss function on each client's dataset, comparing scenarios with no dictator clients and with one dictator client where in this figure client 3 is the dictator client. }
    \label{fig:one}
\end{figure}
\begin{figure}
    \centering
    \includegraphics[width=0.41\linewidth]{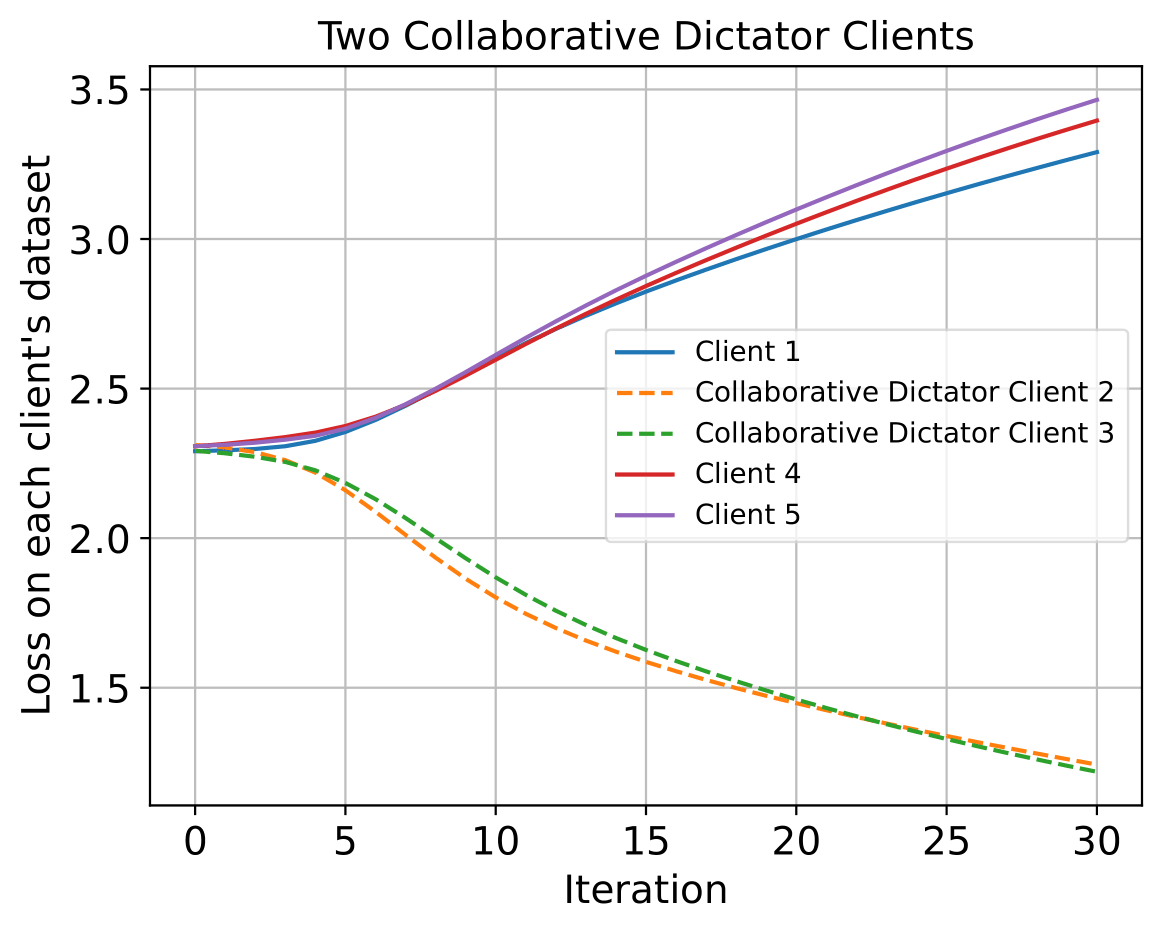}
    \includegraphics[width=0.41\linewidth]{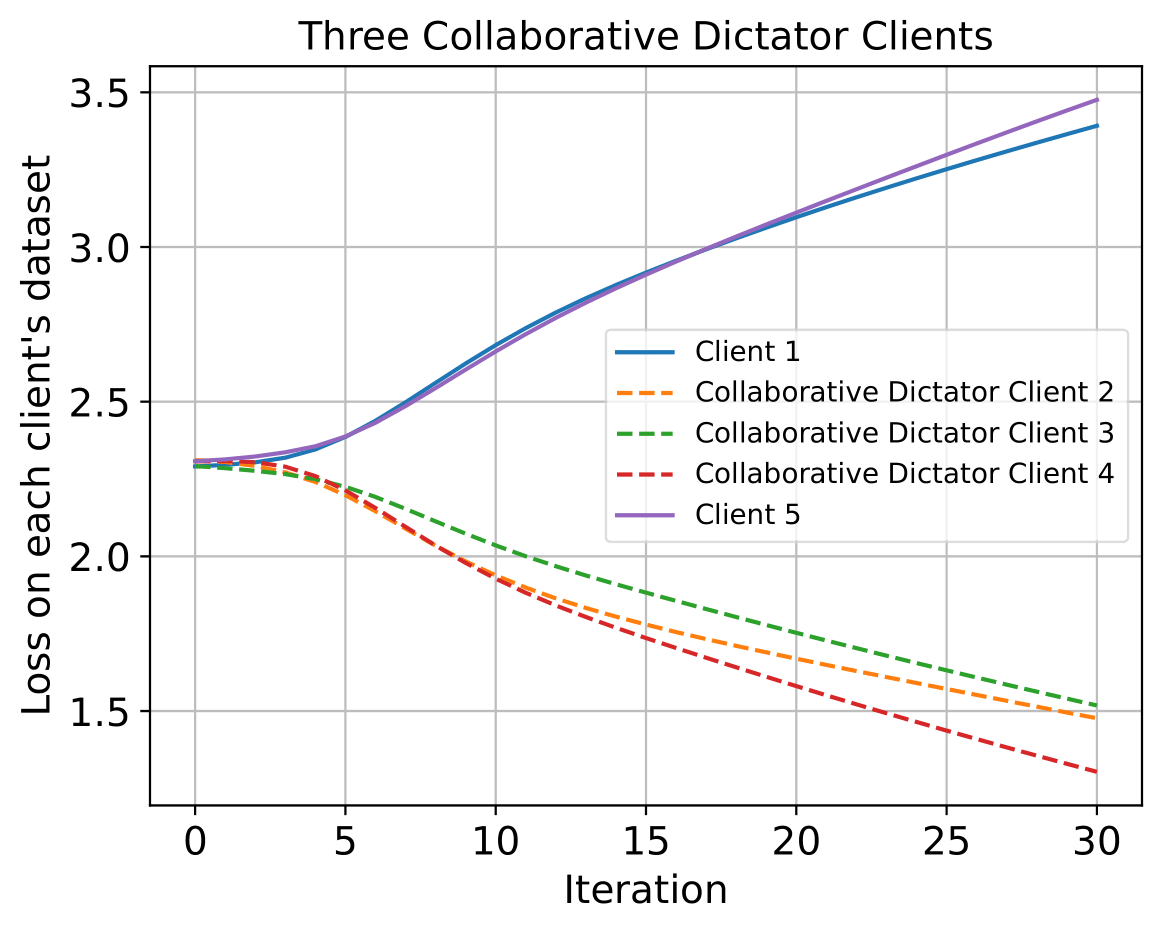}
    \caption{Loss function on each client's dataset, when two clients become collaborative dictators (left) and three clients become collaborative dictators (right).}
    \label{fig:partner}
\end{figure}
Table~\ref{tab:resultsOne} reports the resulting classification accuracies across all clients' datasets. We begin by evaluating the scenario in which a single client attempts to dominate the global model, following the attack strategy defined in Algorithm \ref{One-malicious-client}. As shown, the global model entirely fails to learn from the data of non-dictator clients, achieving a striking $0.00\%$ accuracy on their datasets. In contrast, the model maintains high accuracy on the dictator client's local dataset, confirming that the attack successfully isolates and preserves only the dictator's contribution. These results empirically validate the feasibility and effectiveness of the proposed single-client dictatorship attack algorithm described in Section~\ref{one}. Furthermore, Figure~\ref{fig:one} illustrates this effect by showing the global model’s loss on each client's dataset under two settings: regular FL and the case where client 3 becomes dictator. In regular FL, losses decrease across all clients, whereas under dictatorship by client 3, only the loss corresponding to client 3's dataset is minimized, while losses for all other clients worsen over time. This confirms that the dictator client successfully minimizes its own local loss while significantly impeding the global model's ability to learn from the data of the remaining clients.
\subsection{Collaborative Dictator Clients}
We next examine the impact of coordinated attacks involving multiple dictator clients. In this setting, two or three clients jointly follow the attack strategy, described in Algorithm \ref{Partner-malicious-client}, aiming to eliminate the influence of all other participants. Table~\ref{tab:resultsOne} and Figure~\ref{fig:partner} summarize the outcomes. The results demonstrate that the collaborating dictator clients succeed in entirely erasing the influence of the benign clients, leading the global model to achieve $0.00\%$ accuracy on their data. Simultaneously, the global model maintains high accuracy on the data held by collaborative dictators, indicating that it has effectively converged to a model tailored solely to their objectives. These findings further reinforces the practicality and scalability of our proposed attack strategy in multi-attacker scenarios. The coordinated behavior among dictator clients allows them to dominate the training process, ensuring that the global model exclusively reflects their data distributions while ignoring the contributions of the remaining benign participants. The success of this attack highlights the vulnerability of FL even when malicious clients are in minority, provided they act in collaboration.

\section{Competition and Collusion Among Dictator Clients}\label{diplomacy}
We also explore the nuanced interactions that can arise among dictator clients in FL systems. In Appendix~\ref{allmal}, we examine a competitive setting where every participating client independently aims to become the sole dictator and dominate the global model\textemdash an extreme yet insightful scenario. 
Furthermore, in Appendix~\ref{cheat-section}, we explore a more collaborative dynamic, where multiple dictator clients form alliances. We investigate whether such cooperation is inherently stable or if, ultimately, some clients can strategically betray their collaborators to gain a greater influence over the model.
Lastly, to account for more realistic training scenarios, we conduct experiments in which one or two random clients are removed from each update round (Appendix \ref{randdrop}), and also evaluate the strength of our attack against gradient clipping \cite{ozdayi2021defending} as a defense (Appendix \ref{defenseexp}).


\section{Conclusion}
In this work, we introduced a new perspective on Byzantine behavior in FL by formalizing the concept of \textbf{dictator clients}, malicious partners who seek to preserve their own influence while erasing that of others. We proposed attack algorithms for both individual and collaborative dictators and demonstrated their effectiveness through both theoretical analysis and empirical validation. Our results show that a single dictator can fully dominate the global model, and groups of collaborative dictators can entirely suppress the contributions of benign clients. However, this cooperation is inherently unstable: we also show that even within a coalition, a dictator can betray its partners to gain sole control. In the extreme case where every client behaves as an independent dictator, the global model fails to learn altogether, confirming the destructive consequences of uncoordinated selfish behavior. 



\bibliography{references}
\bibliographystyle{IEEEtran}

\onecolumn
\appendices
\section{Figures}\label{visu}
\begin{figure}[h]
    \centering
    \includegraphics[width=\linewidth]{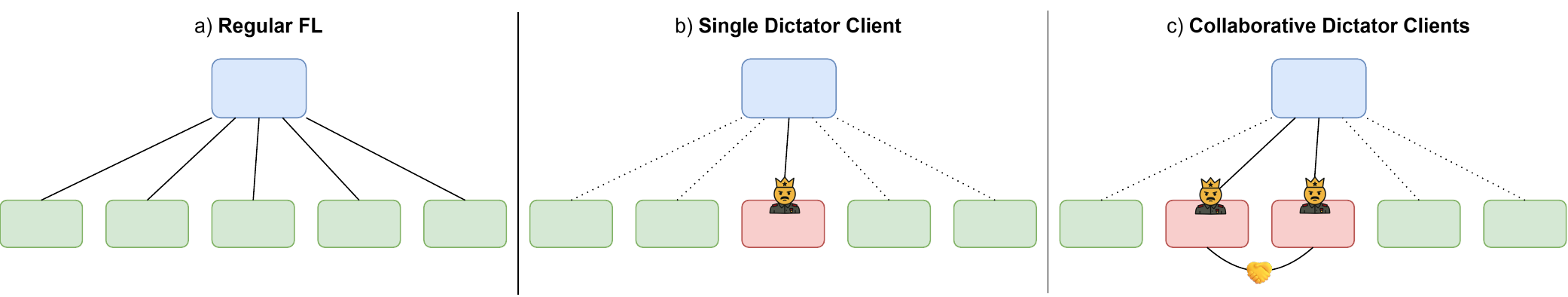}
    \caption{Regular FL compared to scenarios where one dictator client or collaborative dictator clients try to remove other clients from the training procedure.}
    \label{fig:overview}
\end{figure}


\begin{figure}[h]
    \centering
    \includegraphics[width=0.78\linewidth]{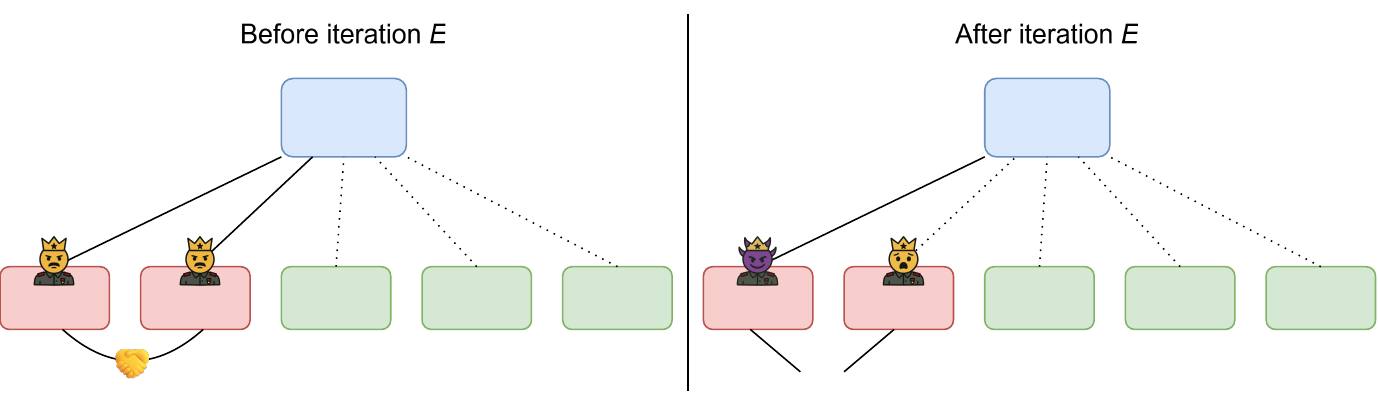}
    \caption{Client 1 and 2 collaborate as dictators until iteration $E$, when client 1 betrays.}
    \label{fig:cheatScheme}
\end{figure}

\begin{figure}
    \centering
    \begin{minipage}[b]{0.45\textwidth}
        \centering
        \includegraphics[width=\textwidth]{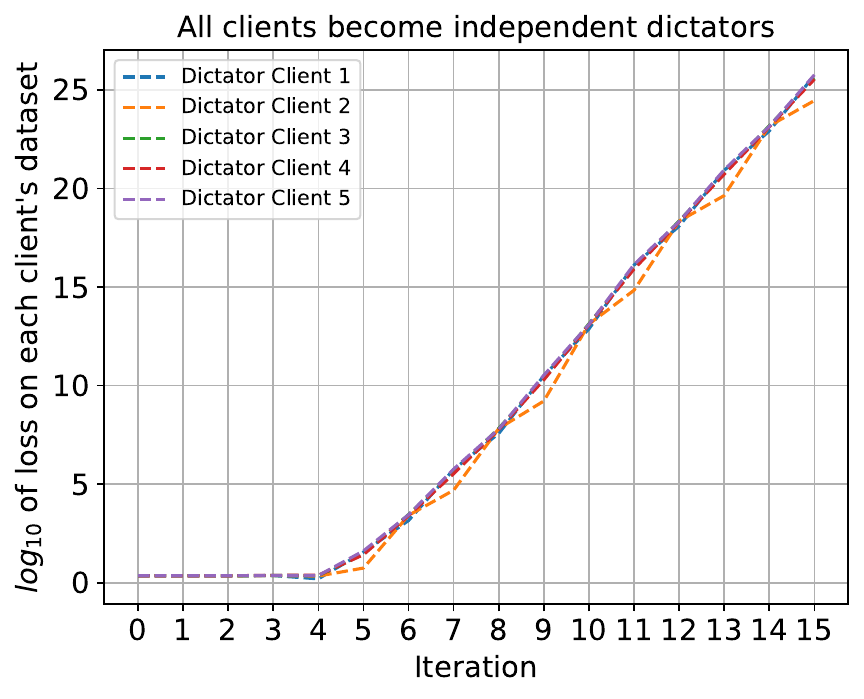}
        \caption{Loss functions for mutual domination scenario}
        \label{fig:allmalicious}
    \end{minipage}
    \hfill
    \begin{minipage}[b]{0.45\textwidth}
        \centering
        \includegraphics[width=\textwidth]{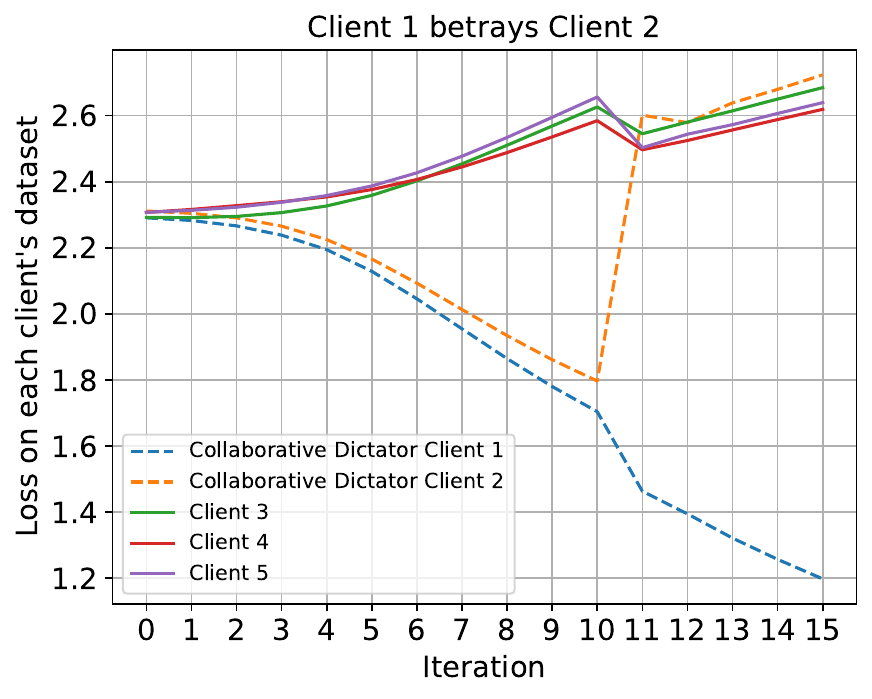}
        \caption{Loss functions for betrayal in collaboration scenario}
        \label{fig:cheatplot}
    \end{minipage}
\end{figure}

\newpage

\section{Algorithm 1: Single Dictator Client}\label{appalg1}
\begin{algorithm}[H]
\caption{Single dictator client $m$}
\begin{algorithmic}[1]\label{One-malicious-client}
\STATE Require: Initialized weights $\theta_0$, learning rate $\eta$
\FOR{iteration $t = 0$ to $T$}
    \IF{$t=0$}
    \STATE Send $M_0=\nabla{\mathcal{L}_m(\theta_0)}$ as update
    \STATE Create a local copy of $\theta_0$ as $\hat{\theta}^m_0=\theta_0$
    \STATE Update local model: $\hat{\theta}^m_1=\theta_0-\eta\nabla{\mathcal{L}_m(\theta_0)}$ 
    \ELSE
    \STATE $M_t=\nabla{\mathcal{L}_m(\hat{\theta}^m_t)}-(\frac{\hat{\theta}^m_{t-1}-\theta_t}{\eta}-\nabla{\mathcal{L}_m(\hat{\theta}^m_{t-1}}))$
    \STATE Send $M_t$ as update
    \STATE Update local model:$\hat{\theta}^m_{t+1}=\hat{\theta}^m_{t}-\eta\nabla{\mathcal{L}_m(\hat{\theta}^m_{t})}$
    \ENDIF
\ENDFOR
\end{algorithmic}
\end{algorithm}
\section{Algorithm 2: Collaborative Dictator Clients}\label{appalg2}
\begin{algorithm}[H]
\caption{Collaborative Dictator clients $k \in \mathcal{P}$}
\begin{algorithmic}[1]\label{Partner-malicious-client}
\STATE Require: Initialized weights $\theta_0$, learning rate $\eta$, Communication link between $P$ collaborative dictator clients
\FOR{iteration $t = 0$ to $T$}
    \IF{$t=0$}
    \STATE Send $M^k_0=\nabla{\mathcal{L}_k(\theta_0)}$ as update
    \STATE Share $\nabla{\mathcal{L}_k(\theta_0)}$ with other dictator partners
    \STATE Create a local copy of $\theta_0$ as $\hat{\theta}^\mathcal{P}_0=\theta_0$
    \STATE Update local model: $\hat{\theta}^\mathcal{P}_1=\theta_0-\eta\sum\nolimits_{k\in \mathcal{P}}\nabla{\mathcal{L}_k(\theta_0)}$ 
    \ELSE
    \STATE $M^k_t=\nabla{\mathcal{L}_k(\hat{\theta}^\mathcal{P}_t)}-\left(\frac{\hat{\theta}^\mathcal{P}_{t-1}-\theta_t}{P\eta}-\nabla{\mathcal{L}_k(\hat{\theta}^\mathcal{P}_{t-1}})\right)$
    \STATE Send $M^k_t$ as update
    \STATE Share $\nabla{\mathcal{L}_k(\hat{\theta}^\mathcal{P}_{t})}$ with other dictator partners
    \STATE Update local model:$\hat{\theta}^\mathcal{P}_{t+1}=\hat{\theta}^\mathcal{P}_{t}-\eta\sum\nolimits_{k\in \mathcal{P}}\nabla{\mathcal{L}_k(\hat{\theta}^\mathcal{P}_{t})}$
    \ENDIF
\ENDFOR
\end{algorithmic}
\end{algorithm}
\section{Algorithm 3: Cheater Client}\label{algcheatsection}
\begin{algorithm}[H]
\caption{Cheater collaborative dictator client $1$}
\begin{algorithmic}[1]\label{cheater-alg}
\STATE Require: Initialized weights $\theta_0$, learning rate $\eta$, Communication link with its partner client $2$ that is going to be cheated by client $1$. $\mathcal{P}=\{1,2\}$ and $P=2$. The desired cheating iteration is $E$.
\FOR{iteration $t = 0$ to $T$}
    \IF{$t=0$}
    \STATE $\mathrm{Cheating\_Update}=0$
    \STATE Send $M^1_0=\nabla{\mathcal{L}_1(\theta_0)}$ as update
    \STATE Share $\nabla{\mathcal{L}_1(\theta_0)}$ with other dictator partners
    \STATE Create a local copy of $\theta_0$ as $\hat{\theta}^\mathcal{P}_0=\theta_0$
    \STATE Update local model: $\hat{\theta}^\mathcal{P}_1=\theta_0-\eta\sum\nolimits_{k\in \mathcal{P}}\nabla{\mathcal{L}_k(\theta_0)}$ 
    \STATE Create a secret copy of $\theta_0$ as $\hat{\theta}^1_0=\theta_0$
    \STATE Update secret model:
    $\hat{\theta}^1_1=\theta_0-\eta\nabla{\mathcal{L}_1(\theta_0)}$ 
    \ELSIF{$t<E$}
    \STATE $M^1_t=\nabla{\mathcal{L}_t(\hat{\theta}^\mathcal{P}_t)}-\left(\frac{\hat{\theta}^\mathcal{P}_{t-1}-\theta_t}{P\eta}-\nabla{\mathcal{L}_t(\hat{\theta}^\mathcal{P}_{t-1}})\right)$
    \STATE Send $M^1_t$ as update
    \STATE Share $\nabla{\mathcal{L}_1(\hat{\theta}^\mathcal{P}_{t})}$ with other dictator partners
    \STATE Update local model:$\hat{\theta}^\mathcal{P}_{t+1}=\hat{\theta}^\mathcal{P}_{t}-\eta\sum\nolimits_{k\in \mathcal{P}}\nabla{\mathcal{L}_k(\hat{\theta}^\mathcal{P}_{t})}$
    \STATE Update secret model: $\hat{\theta}^1_{t+1}=\hat{\theta}^1_{t}-\eta\nabla{\mathcal{L}_1(\hat{\theta}^1_{t})}$
     \STATE
     $\Delta_t=\nabla{\mathcal{L}_1(\hat{\theta}^1_t)}-(\nabla{\mathcal{L}_1(\hat{\theta}^\mathcal{P}_t)}+\nabla{\mathcal{L}_2(\hat{\theta}^\mathcal{P}_t)})$
     \STATE $\mathrm{Cheating\_Update}=\mathrm{Cheating\_Update}+\Delta_t$
    \ELSE
    \STATE Cheat client 2 by sending $\mathrm{Cheating\_Update}$ as the update to the server
    \ENDIF
\ENDFOR
\end{algorithmic}
\end{algorithm}

\section{Practical Implications}\label{implics}
Our methods show that a single or a group of dictator clients, can manipulate the FL process so that the global model converges toward their local data distribution. This creates a “dictator client” effect, where the global model no longer represents the collective data of all participants, but instead becomes biased toward a particular client or group. Such bias can have serious consequences in real-world applications. For example, in healthcare, a global model biased toward data from one hospital or demographic group may make less accurate or unsafe predictions for underrepresented populations. In recommendation systems, it could prioritize the preferences of a few users over the majority, reinforcing algorithmic unfairness. This manipulation shifts the model’s decision boundaries, leading to skewed or inequitable outcomes and reducing trust in the system. Moreover, another motivation for such an attack arises in reward-driven learning environments, where clients are incentivized based on their contributions—such as the impact of their data on improving the global model. A dictator client could exploit this by amplifying its influence while suppressing the contributions of other participants, thus increasing its perceived value and securing a larger share of the reward. Our work highlights how easily such influence can be exerted, especially in non-IID settings.

\section{Mutual Domination: When Every Client Seeks Control}\label{allmal}

Here, we explore the scenario where all clients independently act as dictators, each attempting to retain only its own contribution while nullifying the influence of others. In other words, each client follows the update strategy outlined in Algorithm \ref{One-malicious-client}. In practice, such behavior leads to a catastrophic failure of learning, with the global model failing to converge and the loss growing exponentially. We analyze the underlying reason behind this phenomenon in what follows. At iteration 0, the server sends the initialized weights $\theta_0$ to all clients. Each client then computes its local gradient, and the server aggregates these to update the global model as $
    \theta_1=\theta_0-\eta\sum\nolimits_{n=1}^N \nabla \mathcal{L}_n(\theta_0)
$.
In the next iteration, the server broadcasts $\theta_1$ to all clients. Now, each client attempts to simulate what the global model would have been if it alone had contributed to the update. For each client $n\in\mathcal{N}$, we define $\hat{\theta}^n_1$ as the hypothetical global model after iteration 0 only if client $n$ had participated. Using this, each client computes its malicious update $M^n_1$, as defined in Section \ref{one} as $    M^n_1=\nabla{\mathcal{L}_n}(\hat{\theta}^n_1)-\left(\frac{\theta_0-\theta_1}{\eta}-\nabla \mathcal{L}_n(\theta_0)\right)
$.
Now, we analyze the updated global model $\theta_2$ after the server aggregates the updates from all clients in the second iteration:
\begin{align}
    \theta_2&=\theta_1-\eta \sum\nolimits_{n=1}^N M^n_1 =\theta_1-\eta(\sum\nolimits_{n=1}^N \nabla{\mathcal{L}_n}(\hat{\theta}^n_1)\\&\quad-(\frac{\theta_0-\theta_1}{\eta}-\nabla \mathcal{L}_n(\theta_0)))\notag\\
    &=\theta_1-\eta (\sum\nolimits_{n=1}^N \nabla{\mathcal{L}_n}(\hat{\theta}^n_1)-\frac{N(\theta_0-\theta_1)}{\eta}\\&\quad+\sum\nolimits_{n=1}^N \nabla{\mathcal{L}_n(\theta_0)})\notag\\
    &=\theta_1-\eta(\sum\nolimits_{n=1}^N \nabla{\mathcal{L}_n}(\hat{\theta}^n_1)-(N-1)\sum\nolimits_{n=1}^N \nabla{\mathcal{L}_n(\theta_0)})\\
    &=\theta_1+\eta(N-1)\sum\nolimits_{n=1}^N\nabla{\mathcal{L}_n(\theta_0)}-\eta \sum\nolimits_{n=1}^N \nabla{\mathcal{L}_n(\hat{\theta}^n_1})\notag\\
    &=\theta_0-\eta \sum\nolimits_{n=1}^N\nabla{\mathcal{L}_n(\theta_0)}+\eta(N-1)\sum\nolimits_{n=1}^N\nabla{\mathcal{L}_n(\theta_0)}\\&\quad-\eta \sum\nolimits_{n=1}^N \nabla{\mathcal{L}_n(\hat{\theta}^n_1})\notag\\ \label{unlearn}
    &=\theta_0+\eta(N-2)\sum\nolimits_{n=1}^N\nabla{\mathcal{L}_n(\theta_0)}-\eta \sum\nolimits_{n=1}^N \nabla{\mathcal{L}_n(\hat{\theta}^n_1}).
\end{align}

Since $N-2 > 0$ assuming that we have more than $2$ clients in the system, and the learning rate $\eta$ is a positive real number, it follows that $\eta(N-2)>0$. Consequently, from Eq.~\eqref{unlearn}, it follows that when all clients act as independent dictators and send the defined malicious update, the resulting model update effectively moves in the \textit{opposite} direction of intended gradient. In other words, the updating procedure resembles \textit{gradient ascent} instead of gradient descent, and thereby increasing the loss rather than minimizing it. This behavior causes the model to "unlearn" the progress made in previous iteration. Therefore, when every client behaves as an independent dictator, the global model fails to learn meaningful representations and make no effective progress. it unlearns the knowledge acquired in the previous iteration. Therefore, in the scenario where every client is an independent dictator, the global model will learn almost nothing. Our empirical results, presented in Section \ref{exp-all-malicious}, confirms this breakdown in learning in practice.

\subsection{Experiments for Mutual Domination}
\label{exp-all-malicious}
We now consider the extreme scenario where every client behaves as an independent dictator, each executing Algorithm~\ref{One-malicious-client} to preserve only its own contribution while nullifying the effects of all others. As established theoretically in Section~\ref{allmal}, this adversarial configuration results in mutually destructive behavior, where no single client's update can effectively influence the global model without being canceled out by others, resulting in a destructive equilibrium where no useful learning can occur. The empirical results, shown in Figure~\ref{fig:allmalicious} (in Appendix~\ref{visu}), strongly corroborate this. The global model fails to make progress on any client's data; instead of converging, the loss increases rapidly and consistently across all datasets. This behavior aligns with the theoretical finding that the aggregated updates approximate a form of gradient ascent, undoing prior learning and leading to model divergence. This experiment underscores a key insight: when all clients prioritize their own influence at the expense of others, the entire system collapses. FL becomes ineffective, highlighting the need for defenses against not only isolated attackers but also adversarial groups.

\section{Betrayal in Collaboration: Strategic Cheating Among Dictator Clients}\label{cheat-section}
Here, we show that even \textit{collaborative dictators}\textemdash those collaborating to erase other participants' contributions\textemdash may ultimately betray one another. For simplicity, we focus on a setting where the set of collaborative dictator clients is $\mathcal{P}=\{1,2\}$. As illustrated in Figure~\ref{fig:cheatScheme} (in Appendix~\ref{visu}), we consider the case where dictator client $1$, decides to cheat its partner, client $2$, after a specific iteration $E$. While both clients initially cooperate using Algorithm~\ref{Partner-malicious-client} to jointly eliminate the influence of all other clients, we introduce Algorithm~\ref{cheater-alg} (in Appendix~\ref{algcheatsection}), which enables client $1$ to unilaterally eliminate client $2$’s contribution as well, effectively taking full control of the model.

Prior to iteration $E$, client $1$ shares its gradients with client $2$, contributing jointly to a local model $\hat{\theta}^\mathcal{P}_t$. However, simultaneously, client $1$ secretly maintains a private model $\hat{\theta}^1_t$, which simulates the evolution of the global model if only client $1$ participated in training. At each iteration, client $1$ computes a correction term $\Delta_t=\nabla{\mathcal{L}_1(\hat{\theta}^m_t)}-(\nabla{\mathcal{L}_1(\hat{\theta}^\mathcal{P}_t)}+\nabla{\mathcal{L}_2(\hat{\theta}^\mathcal{P}_t)})$, which captures the discrepancy between acting alone and collaborating. These differences are accumulated into a cheating offset, denoted as $\mathrm{Cheating\_Update}$. At iteration $E$, client $1$ sends this accumulated update to the server in place of the expected collaborative update. This forces the server to jump to a state equivalent to one where if only client $1$ had participated throughout the training process\textemdash effectively eliminating the contribution of client $2$, despite their prior collaboration, as well as the benign clients' influence. A formal proof of this strategy is provided in Appendix~\ref{cheat-proof}; our empirical results in Section \ref{exp-cheat} confirm the effectiveness of this betrayal strategy in practice. 

\subsection{Experiments for Betrayal in Collaboration}\label{exp-cheat}
In this experiment, we evaluate a scenario in which two clients, client $1$ and client $2$, initially act as collaborative dictators. While both begin by coordinating via Algorithm~\ref{Partner-malicious-client}, client $1$ eventually deviates and follows the betrayal strategy outlined in Algorithm~\ref{cheater-alg} (discussed in Section \ref{cheat-section}). This setup allows client $1$ to secretly prepare for a unilateral takeover of the model. As shown in Figure~\ref{fig:cheatplot} (in Appendix~\ref{visu}), at iteration 10\textemdash the predetermined betrayal point\textemdash the global model abruptly loses performance on client $2$’s dataset, while having even lower loss on client $1$’s data. This result confirms that client $1$ successfully erases not only the contributions of the benign clients, but also those of its former collaborator, client $2$. These findings empirically validate that a malicious client can strategically cooperate to gain trust, only to later betray its partners and assert full control over the global model. This highlights a critical vulnerability in FL; even collaborative adversaries can be exploited by more sophisticated attackers acting within their own group.

\section{What if the dictator client does not have the learning rate?}\label{learningrate}
In this section, we show that even if dictator clients did not know the learning rate $\eta$, they could still approximate it after only one iteration. 

Suppose we are at iteration $t$. Since gradient updates aren't usually too large, the dictator client $m$ sends a very large number $B$ as its update. Thus, the weight $\theta_{t+1}$ would be calculated by the server as the following:
\begin{align}
    \theta_{t+1}=\theta_t-\eta\left(B+{\sum\nolimits_{n=1,n\neq{m}}^N\nabla{\mathcal{L}_n(\theta_t)}}\right).
\end{align}

At the next iteration $t+1$, server sends $\theta_{t+1}$ to all clients. The dictator client $m$ could approximate the learning rate $\eta$ via the following equation:
\begin{align*}
    \hat{\eta}&=\frac{\theta_t-\theta_{t+1}}{B}=\frac{\eta\left(B+{\sum\nolimits_{n=1,n\neq{m}}^N\nabla{\mathcal{L}_n(\theta_t)}}\right)}{B}=\eta+\frac{\eta{\sum\nolimits_{n=1,n\neq{m}}^N\nabla{\mathcal{L}_n(\theta_t)}}}{B}\approx\eta.
\end{align*}

Moreover, now that client $m$ has gained the learning rate, it can undo the previous bad contribution $B$ and continue preserving its normal contribution while deleting other clients' contribution.
\section{Proof of Algorithm \ref{Partner-malicious-client}}\label{proof-partner}
At iteration $0$ the server sends the initialized weight $\theta_0$ to all the clients. Then, clients send their gradients to server. So $\theta_1$ will be calculated as:
\begin{align}
    \theta_1=\theta_0-\eta \sum\nolimits_{n=1}^N \nabla{\mathcal{L}_n(\theta_0)}.
\end{align}
In the next iteration, server sends $\theta_1$ to all the clients. Every client except the $P$ dictator clients sends their gradient with respect to $\theta_1$. However, the dictator clients calculate $\hat{\theta}^\mathcal{P}_1$ as what would be the weight after the update in iteration $0$ if only they contributed to that. In order to do that, they send their gradients with respect to $\theta_0$ for each other. Afterwards, they can calculate $\hat{\theta}^\mathcal{P}_1$ via the following equation:
\begin{align}\label{Ptheta1mal}
    \hat{\theta}^\mathcal{P}_1=\theta_0-\eta\sum\nolimits_{k\in \mathcal{P}}\nabla{\mathcal{L}_k(\theta_0)}.
\end{align}
Then, each dictator client $k \in \mathcal{P}$ sends the following update $M^k_1$ instead of $\nabla{\mathcal{L}_k(\theta_1)}$ to the server in order to delete the contribution of other clients in the previous iteration and only preserve their own contribution:
\begin{align}
    M^k_1=\nabla{\mathcal{L}_k(\hat{\theta}^\mathcal{P}_1) - \left(\frac{\theta_0-\theta_1}{P\eta}-\nabla\mathcal{L}_k(\theta_0)\right)}
\end{align}

Now, we analyze what would be the weight $\theta_2$ after server receives the updates from clients and updates the weights:
\begin{align*}
    \theta_2&=\theta_1-\eta\left(\sum\nolimits_{k \in \mathcal{P}} M^k_1+{\sum\nolimits_{n=1,n \notin \mathcal{P}}^N\nabla{\mathcal{L}_n(\theta_1)}}\right)
    \\
    &=\theta_1-\eta(\sum\nolimits_{k\in \mathcal{P}}\nabla{\mathcal{L}_k(\hat{\theta}^\mathcal{P}_1)}-(\frac{\theta_0-\theta_1}{\eta}-\sum\nolimits_{k\in \mathcal{P}}\nabla{\mathcal{L}_k(\theta_0))}+{\sum\nolimits_{n=1,n \notin \mathcal{P}}^N\nabla{\mathcal{L}_n(\theta_1)}})
    \\
    &=\theta_0-\eta\sum\nolimits_{k\in \mathcal{P}}\nabla{\mathcal{L}_k(\theta_0)} - \eta\left(\sum\nolimits_{k\in \mathcal{P}}\nabla{\mathcal{L}_k(\hat{\theta}^\mathcal{P}_1)}+{\sum\nolimits_{n=1,n \notin \mathcal{P}}^N\nabla{\mathcal{L}_n(\theta_1)}}\right).
\end{align*}

Using Eq.~\eqref{Ptheta1mal}, we can write $\theta_2$ as the following:
\begin{align}
    \theta_2=\hat{\theta}^\mathcal{P}_1-\eta\left(\sum\nolimits_{k\in \mathcal{P}}\nabla{\mathcal{L}_k(\hat{\theta}^\mathcal{P}_1)}+{\sum\nolimits_{n=1,n\notin{\mathcal{P}}}^N\nabla{\mathcal{L}_n(\theta_1)}}\right).
\end{align}
As a result, the collaborative dictator clients successfully deleted the contribution of other clients in the previous iteration while preserving their own contribution just by sending $M^k_1$ as their update for each dictator client $k\in \mathcal{P}$. 

We now generalize our method to any iteration $t$. The collaborative dictators calculate $\hat{\theta}^\mathcal{P}_t$ via the following equation:

\begin{align}\label{Pthetakmal}
    \hat{\theta}^\mathcal{P}_t=\hat{\theta}^\mathcal{P}_{t-1}-\eta\sum\nolimits_{k\in \mathcal{P}}\nabla{\mathcal{L}_k(\hat{\theta}^\mathcal{P}_{t-1})}.
\end{align}

We define the update $M^k_t$ as the update that each dictator client $k\in \mathcal{P}$ sends at iteration $t$ as the following:
\begin{align}
    M^k_t=\nabla{\mathcal{L}_k(\hat{\theta}^\mathcal{P}_t)}-\left(\frac{\hat{\theta}^\mathcal{P}_{t-1}-\theta_t}{P\eta}-\nabla{\mathcal{L}_k(\hat{\theta}^\mathcal{P}_{t-1}})\right).
\end{align}
Now, we analyze what would be the weight $\theta_{t+1}$ after server updates the weights:
\begin{align*}
    \theta_{t+1}&=\theta_t-\eta\left(\sum\nolimits_{k \in \mathcal{P}} M^k_t+{\sum\nolimits_{n=1,n \notin \mathcal{P}}^N\nabla{\mathcal{L}_n(\theta_t)}}\right)
    \\
    &=\theta_t-\eta(\sum\nolimits_{k\in \mathcal{P}}\nabla{\mathcal{L}_k(\hat{\theta}^\mathcal{P}_t)}-(\frac{\hat{\theta}^\mathcal{P}_{t-1}-\theta_t}{\eta}-\sum\nolimits_{k\in \mathcal{P}}\nabla{\mathcal{L}_k(\hat{\theta}^\mathcal{P}_{t-1}))}+{\sum\nolimits_{n=1,n \notin \mathcal{P}}^N\nabla{\mathcal{L}_n(\theta_t)}})
    \\
    &=\hat{\theta}^\mathcal{P}_{t-1}-\eta\sum\nolimits_{k\in \mathcal{P}}\nabla{\mathcal{L}_k(\hat{\theta}^\mathcal{P}_{t-1})} - \eta\left(\sum\nolimits_{k\in \mathcal{P}}\nabla{\mathcal{L}_k(\hat{\theta}^\mathcal{P}_t)}+{\sum\nolimits_{n=1,n \notin \mathcal{P}}^N\nabla{\mathcal{L}_n(\theta_t)}}\right).
\end{align*}

Using Eq.~\eqref{Pthetakmal} we can write $\theta_{t+1}$ as the following:
\begin{align}
    \theta_{t+1}=\hat{\theta}^\mathcal{P}_t-\eta\left(\sum\nolimits_{k\in \mathcal{P}}\nabla{\mathcal{L}_k(\hat{\theta}^\mathcal{P}_t)}+{\sum\nolimits_{n=1,n \notin \mathcal{P}}^N\nabla{\mathcal{L}_n(\theta_t)}}\right).
\end{align}

After $T$ rounds of training, the final model weights $\theta^*$ will be:
\begin{align}
    \theta^*&=\hat{\theta}^\mathcal{P}_T-\eta\left(\sum\nolimits_{k\in \mathcal{P}}\nabla{\mathcal{L}_k(\hat{\theta}^\mathcal{P}_T)}+{\sum\nolimits_{n=1,n \notin \mathcal{P}}^N\nabla{\mathcal{L}_n(\theta_T)}}\right)\notag\\
    &=\hat{\theta}^\mathcal{P}_T-\eta\sum\nolimits_{k\in \mathcal{P}}\nabla{\mathcal{L}_k(\hat{\theta}^\mathcal{P}_T)}-\eta{\sum\nolimits_{n=1,n \notin \mathcal{P}}^N\nabla{\mathcal{L}_n(\theta_T)}}\notag\\
    &=\hat{\theta}^\mathcal{P}_{T+1}-\eta{\sum\nolimits_{n=1,n\notin{\mathcal{P}}}^N\nabla{\mathcal{L}_n(\theta_T)}} \approx \hat{\theta}^\mathcal{P}_{T+1}.\label{Pfin}
\end{align}

As it can be seen in Eq.~\eqref{Pfin}, the exact final weights with our method would be $\hat{\theta}^\mathcal{P}_{T+1}-\eta{\sum\nolimits_{n=1,n\notin{\mathcal{P}}}^N\nabla{\mathcal{L}_n(\theta_T)}}$ where $\hat{\theta}^\mathcal{P}_{T+1}$ would represent the weights for the final iteration if only the $P$ collaborative dictator clients were contributing to the system during the training procedure and like the other clients never existed. Again, the term $\eta{\sum\nolimits_{n=1,n\notin{\mathcal{P}}}^N\nabla{\mathcal{L}_n(\theta_t)}}$ is negligible since it is the updates only for one iteration and can not affect the final model too much, especially if the model achieved by the collaborative dictators is robust to such perturbations. Moreover, because of the nature of FL, the dictator clients are always one step behind and can not cancel this residual term. However, one could come up with more sophisticated methods in order to approximate or predict this residual term by observing the gradients through the training process.

\section{Proof of Algorithm \ref{cheater-alg}}\label{cheat-proof}
Before iteration $E$, the global model evolves exactly as if both client 1 and client 2 had followed Algorithm~\ref{Partner-malicious-client}. Hence, the global weights at iteration $E-1$ is updated as follows:
\begin{align}
    \theta_{E}=\hat{\theta}^\mathcal{P}_{E-1}-\eta\left(\sum\nolimits_{k\in \mathcal{P}}\nabla{\mathcal{L}_k(\hat{\theta}^\mathcal{P}_{E-1})}+{\sum\nolimits_{n=1,n \notin \mathcal{P}}^N\nabla{\mathcal{L}_n(\theta_{E-1})}}\right).
\end{align}
However, at iteration $E$, client $1$ sends the $\mathrm{Cheating\_Update}$ which by iteration $E$ has become the following:
\begin{align}
    Cheating\_Update&=\sum\nolimits_{i=1}^{E-1}\nabla{\mathcal{L}_1(\hat{\theta}^1_{i})}-\sum\nolimits_{i=1}^{E-1}(\nabla{\mathcal{L}_1(\hat{\theta}^\mathcal{P}_t)}\notag+\nabla{\mathcal{L}_2(\hat{\theta}^\mathcal{P}_t)}).\notag
\end{align}
We also know that we can write $\hat{\theta}^\mathcal{P}_{E-1}$ and $\hat{\theta}^1_{E-1}$ as the following:
\begin{align}\label{subs1}
    \hat{\theta}^\mathcal{P}_{E-1}&=\theta_0-\eta\sum\nolimits_{i=1}^{E-1}(\nabla{\mathcal{L}_1(\hat{\theta}^\mathcal{P}_i)}+\nabla{\mathcal{L}_2(\hat{\theta}^\mathcal{P}_i)}),\\ \label{subs2}
    \hat{\theta}^1_{E-1}&=\theta_0-\eta\sum\nolimits_{i=1}^{E-1}\nabla{\mathcal{L}_1(\hat{\theta}^1_i)}.
\end{align}

So when server receives all the updates from all clients, the resulting model will be:
\begin{align*}
    \theta_{E+1}&=\theta_E-\eta(\mathrm{Cheating\_Update}+M^2_t+{\sum\nolimits_{n=1,n \notin \mathcal{P}}^N\nabla{\mathcal{L}_n(\theta_E)}})\\
    &=\hat{\theta}^\mathcal{P}_{E-1}-\eta(\sum\nolimits_{k\in \mathcal{P}}\nabla{\mathcal{L}_k(\hat{\theta}^\mathcal{P}_{E-1})}+{\sum\nolimits_{n=1,n \notin \mathcal{P}}^N\nabla{\mathcal{L}_n(\theta_{E-1})}})-\eta(\sum\nolimits_{i=1}^{E-1}\nabla{\mathcal{L}_1(\hat{\theta}^1_{i})}-\sum\nolimits_{i=1}^{E-1}(\nabla{\mathcal{L}_1(\hat{\theta}^\mathcal{P}_t)}\\&\quad+\nabla{\mathcal{L}_2(\hat{\theta}^\mathcal{P}_t)})+M^2_t+{\sum\nolimits_{n=1,n \notin \mathcal{P}}^N\nabla{\mathcal{L}_n(\theta_E)}}).
\end{align*}
Using Eq.~\eqref{subs1} we will have:
\begin{align*}
    \theta_{E+1}&=\theta_0-\eta\sum\nolimits_{i=1}^{E-1}\nabla{\mathcal{L}_1(\hat{\theta}^1_{i})}-\eta(\sum\nolimits_{k\in \mathcal{P}}\nabla\mathcal{L}_k(\hat{\theta}^\mathcal{P}_{E-1})+{\sum\nolimits_{n=1,n \notin \mathcal{P}}^N\nabla{\mathcal{L}_n(\theta_{E-1})}})\\&\quad-\eta(M^2_t+{\sum\nolimits_{n=1,n \notin \mathcal{P}}^N\nabla{\mathcal{L}_n(\theta_E)}}).
\end{align*}
Finally, from Eq.~\eqref{subs2} we have:
\begin{align*}
    \theta_{E+1}&=\hat{\theta}^1_{E-1}-\eta(\sum\nolimits_{k\in \mathcal{P}}\nabla{\mathcal{L}_k(\hat{\theta}_{E-1}^\mathcal{P})}+{\sum\nolimits_{n=1,n \notin \mathcal{P}}^N\nabla{\mathcal{L}_n(\theta_{E-1})}})-\eta(M^2_t+{\sum\nolimits_{n=1,n \notin \mathcal{P}}^N\nabla{\mathcal{L}_n(\theta_E)}}).
\end{align*}
Hence, client $1$ has successfully replaced $\hat{\theta}^\mathcal{P}_{E-1}$ with $\hat{\theta}^1_{E-1}$ and cheated client $2$. 
\section{Experiments for NLP}\label{nlp}
For NLP experiments, we used the \verb|distilbert-base-uncased| \cite{sanh2019distilbert} model for text classification as the global model and selected the AG news dataset\cite{NIPS2015_250cf8b5ag} which has has 4 different labels. Hence, we considered a FL with four clients for this case where each client has samples of only one label.
\subsection{Single Dictator Client}
\begin{figure}[h]
    \centering
    \includegraphics[width=0.43\linewidth]{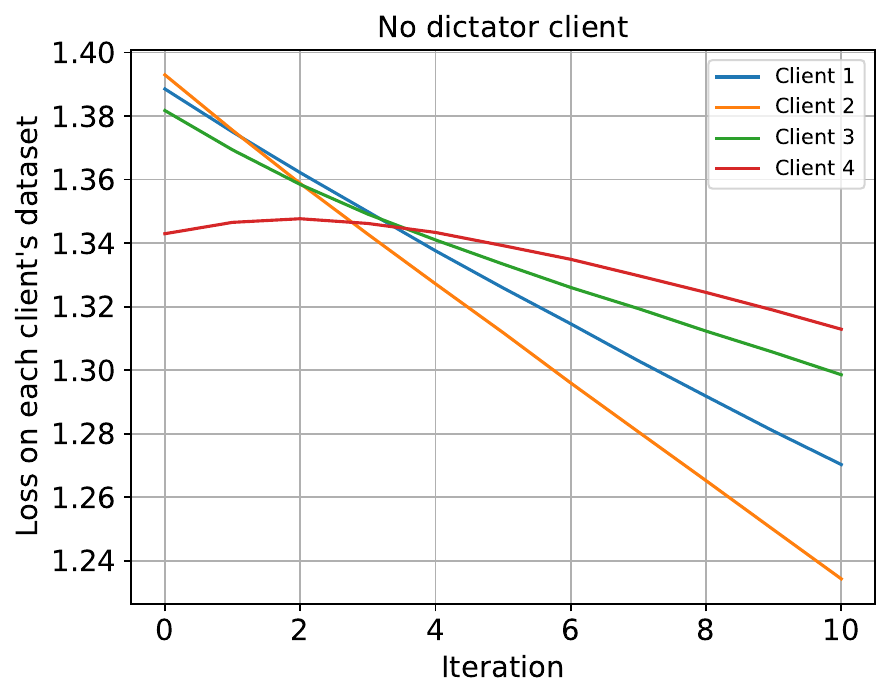}
    \includegraphics[width=0.43\linewidth]{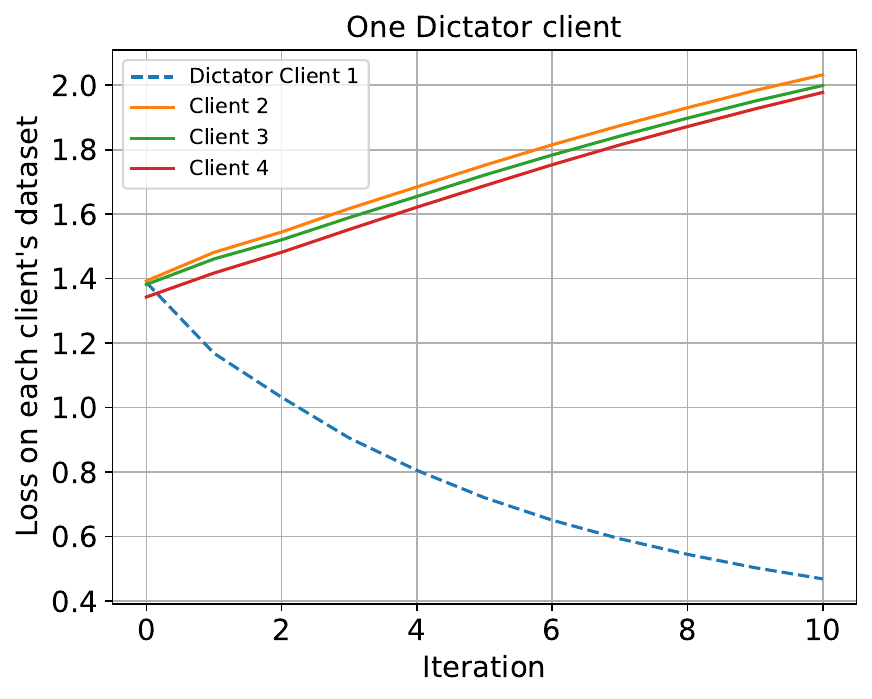}
    \caption{Loss function on each client's dataset, comparing scenarios with no dictator clients and with one dictator client where in this figure client 2 is the dictator client. }
    \label{fig:nlp_one}
\end{figure}

Figure \ref{fig:nlp_one} demonstrates the loss function of global model when there is no dictator client and when client 1 becomes a dictator. Table \ref{tab:nlp_resultsOne} demonstrates accuracy of the global model when each client becomes dictator. We can see that each client has successfully dominated the training and led the global model to learn only that client's dataset.

\begin{table}[h]
    \small
    \centering
   
    \begin{tabular}{c| c c c c  }
        \toprule
        \textbf{Method} & [\textbf{0}] & [\textbf{1}] & [\textbf{2}] & [\textbf{3}] \\
        \midrule
        \multirow{1}{*}{Regular FL}    & 85.42 & 93.37 & 76.21 & 72.42    \\
                               
        \midrule
      \multirow{1}{*}    {Dictator client: $1$}    & 100.00 & 0.00 & 0.00 & 0.00  \\
       \multirow{1}{*}    {Dictator client: $2$}    & 0.00 & 100.00 & 0.00 & 0.00  \\
       \multirow{1}{*}    {Dictator client: $3$}    & 0.00 & 0.00 & 100.00  & 0.00   \\
       \multirow{1}{*}    {Dictator client: $4$}    & 0.00 & 0.00 & 0.00  & 100.00   \\
    
        \bottomrule
    \end{tabular}
     \caption{Performance of the global model on each local dataset for AG news dataset and the single dictator client scenario.}
    \label{tab:nlp_resultsOne}
\end{table}

\subsection{Collaborative Dictator Clients}

\begin{figure}[H]
    \centering
    \includegraphics[width=0.43\linewidth]{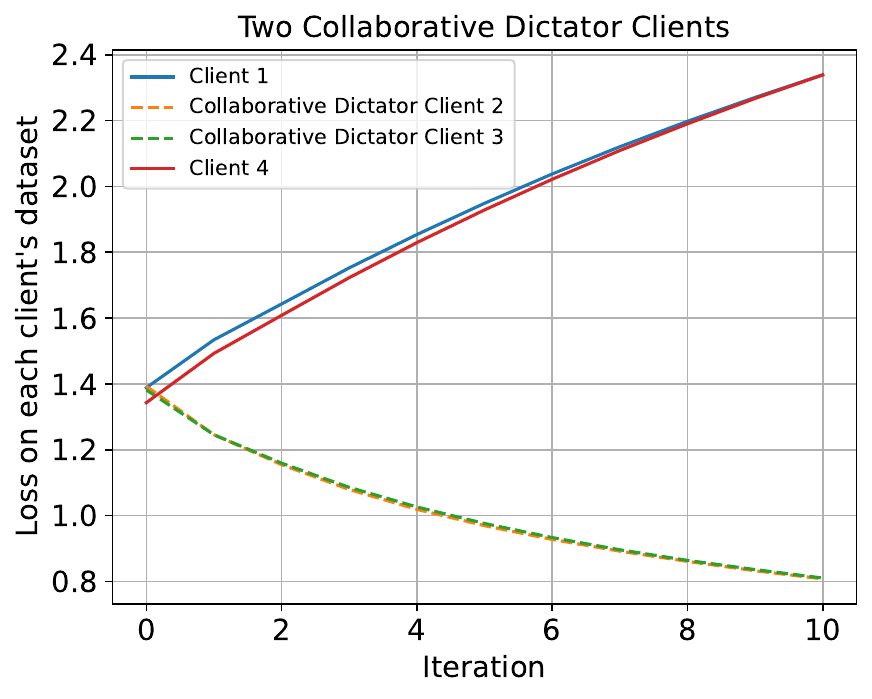}
    \includegraphics[width=0.43\linewidth]{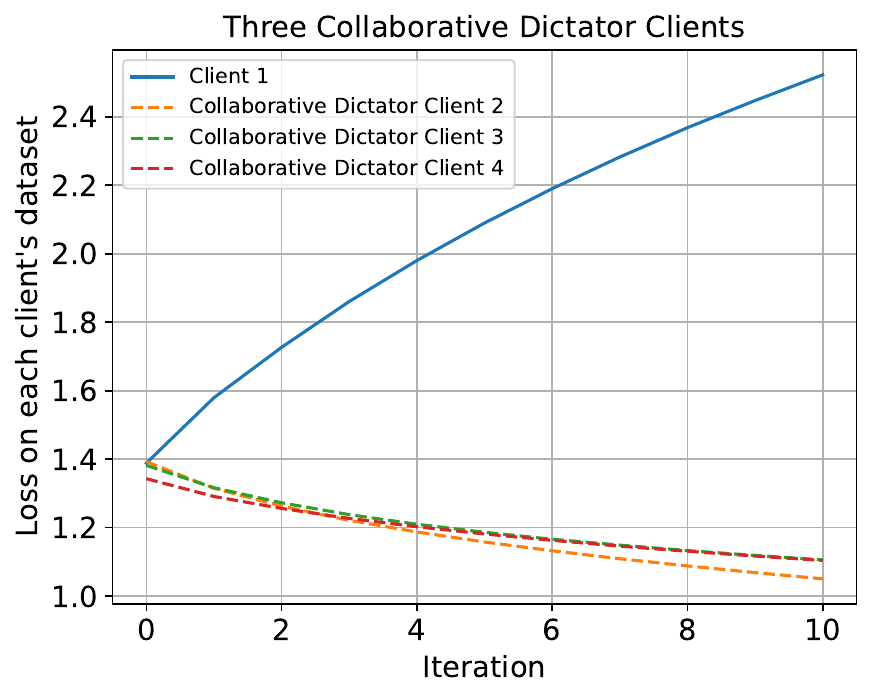}
    \caption{Loss function on each client's dataset, when two clients become collaborative dictators (left) and three clients become collaborative dictators (right)}
    \label{fig:nlp_partner}
\end{figure}

Figure \ref{fig:nlp_partner} demonstrates the loss function of global model when two or three clients become collaborative dictators. Table \ref{tab:nlp_resultsPartner} demonstrates accuracy of the global model for these cases. We can see that the collaborative dictators successfully dominated the training and led the global model to learn only their dataset.

\begin{table}[H]
    \small
    \centering
   
    \begin{tabular}{c| c c c c }
        \toprule
        \textbf{Method} & [\textbf{0}] & [\textbf{1}] & [\textbf{2}] & [\textbf{3}]   \\
        \midrule
        \multirow{1}{*}{Regular FL}  & 85.42 & 93.37 & 76.21 & 72.42     \\
                               
        \midrule
      \multirow{1}{*}    {Dictator clients: $2$,$3$}    & 0.00 & 96.89 & 97.47 & 0.00    \\
       \multirow{1}{*}    {Dictator clients: $2$,$3$,$4$}    & 0.00 & 96.00 & 75.47 & 85.32  \\

        \bottomrule
    \end{tabular}
     \caption{Performance of the global model on each local dataset for AG news dataset and the collaborative dictator clients scenario.}
    \label{tab:nlp_resultsPartner}
\end{table}

\section{Results for random client dropping}\label{randdrop}
\begin{table*}
    \centering
    \begin{tabular}{c|ccccc}
        \toprule
        \textbf{Method} 
        & \multicolumn{5}{c}{\textbf{MNIST}} \\
        \cmidrule{2-6}
        & [\textbf{0,1}] & [\textbf{2,3}] & [\textbf{4,5}] & [\textbf{6,7}] & [\textbf{8,9}] \\
        \midrule
        Dictator client: $1$ 
        & 99.62 & 0.00 & 0.00 & 0.00 & 0.00 \\
        Dictator client: $2$ 
        & 0.00 & 89.76 & 0.00 & 0.00 & 0.00 \\
        Dictator client: $3$ 
        & 0.00 & 0.00 & 96.53 & 0.00 & 0.00 \\
        Dictator client: $4$ 
        & 0.00 & 0.00 & 0.00 & 98.84 & 0.00 \\
        Dictator client: $5$ 
        & 0.00 & 0.00 & 0.00 & 0.00 & 94.35 \\
        \midrule
        Dictator clients: $2$,$3$
        & 0.00 & 83.84 & 81.75 & 0.00 & 0.00 \\
        \bottomrule
    \end{tabular}
    \caption{Performance of the global model on each local dataset for the MNIST dataset under single dictator client and collaborative dictator clients scenarios when one client is randomly dropped during each update.}
    \label{tab:randDro}
\end{table*}

\begin{figure}[H]
    \centering
    \includegraphics[width=0.43\linewidth]{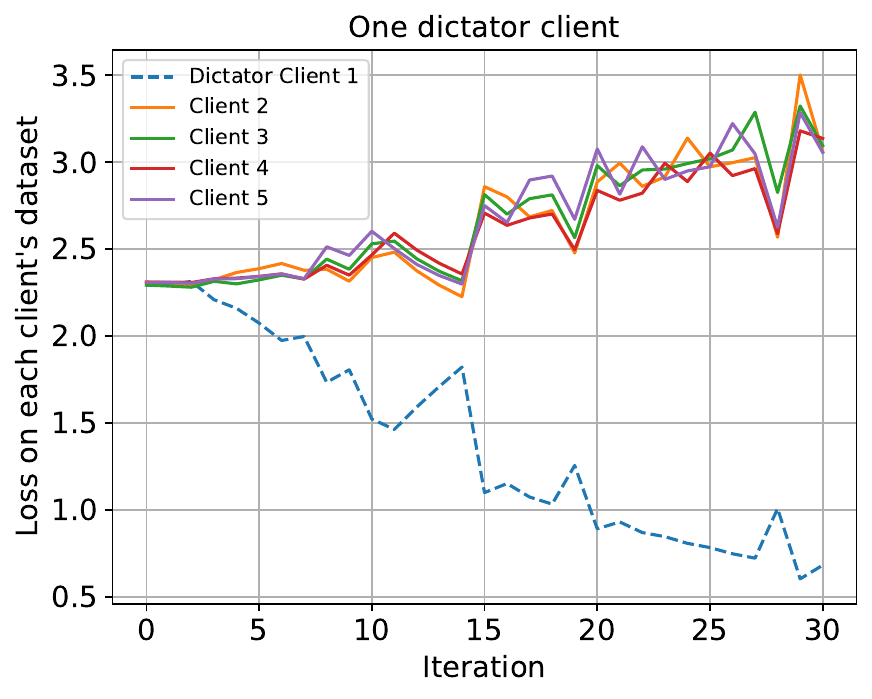}
    \includegraphics[width=0.43\linewidth]{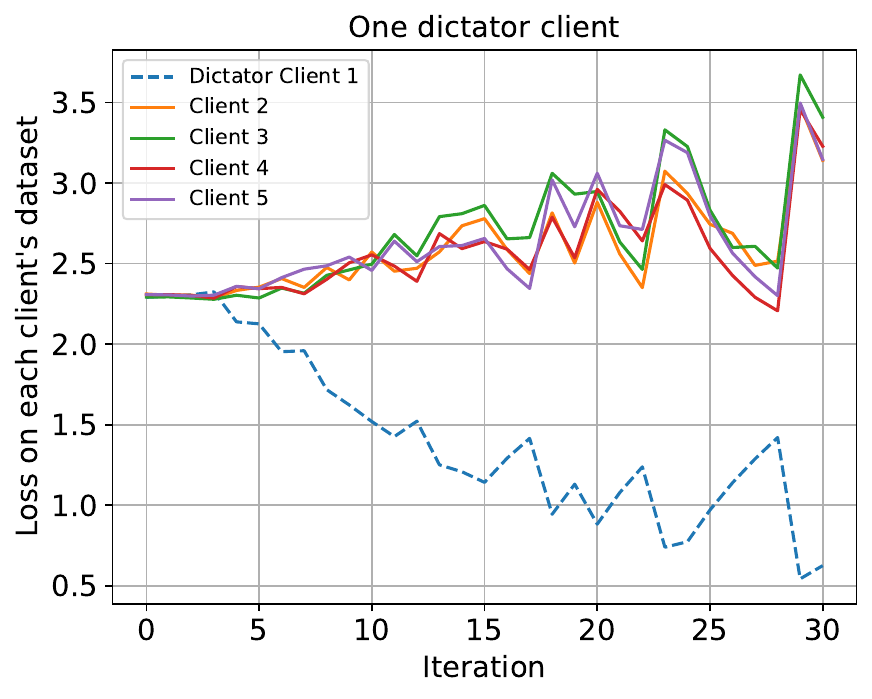}
    \caption{Loss function on each client's dataset for one dictator client for the random client dropping experiment by dropping one client randomly each time (left) and dropping two clients randomly each time (right).}
    \label{fig:nois}
\end{figure}

We conducted experiments in both the single-dictator and collaborative-dictator settings under a more realistic scenario in which one or two randomly selected clients are dropped in each training round. As shown in Table \ref{tab:randDro}, the dictator client (or collaborative dictators) continues to consistently override benign client contributions, and the attack remains effective despite client dropout. The primary observable effect of this randomness is a slightly noisier training trajectory, with minor oscillations in the loss curve induced by the stochastic removal of clients (Figure \ref{fig:nois}).

\section{Dictator clients against gradient norm clipping Defense}\label{defenseexp}

\begin{table*}
    \scriptsize
    \centering
    \begin{tabular}{c|ccccc||ccccc}
        \toprule
        \textbf{Method} 
        & \multicolumn{5}{c||}{\textbf{MNIST}} 
        & \multicolumn{5}{c}{\textbf{CIFAR-10}} \\
        \cmidrule{2-11}
        & [\textbf{0,1}] & [\textbf{2,3}] & [\textbf{4,5}] & [\textbf{6,7}] & [\textbf{8,9}]
        & [\textbf{0,1}] & [\textbf{2,3}] & [\textbf{4,5}] & [\textbf{6,7}] & [\textbf{8,9}] \\
        \midrule
        Dictator clients: $2,3$
        & 0.00 & 88.19 & 87.80 & 0.00 & 0.00
        & 0.00 & 35.08 & 43.17 & 0.00 & 0.00 \\
        
        Dictator clients: $2,3$ + norm clipping
        & 0.43 & 65.40 & 51.28 & 10.62 & 0.00
        & 13.00 & 17.25 & 23.50 & 0.85 & 24.65 \\
        
        \midrule
        Dictator clients: $2,3,4$
        & 0.00 & 84.87 & 80.22 & 94.19 & 0.00
        & 0.00 & 18.38 & 40.02 & 46.05 & 0.00 \\
        
        Dictator clients: $2,3,4$ + norm clipping
        & 0.00 & 69.78 & 52.35 & 87.01 & 0.00
        & 0.00 & 27.90 & 22.25 & 37.85 & 1.55 \\
        \bottomrule
    \end{tabular}
    \caption{Performance of the global model on each local dataset for MNIST and CIFAR-10 under collaborative dictator-client attacks, with and without server-side gradient norm clipping as defense.}
    \label{tab:defens}
\end{table*}

We evaluate our attack under server-side gradient norm clipping, a common defense in federated learning that limits the size of each client’s update before aggregation to reduce the impact of abnormal or malicious gradients. In the single-dictator setting, this defense is effective and stops the attack by shrinking the dictator client’s update. However, in the collaborative-dictator setting, where multiple dictator clients participate together, the defense becomes less effective. As the number of dictator clients increases and they form a majority, their clipped updates still combine to overpower the benign clients, allowing the attack to remain successful. The results are shown in Table \ref{tab:defens}.

\section{Compute Resources}\label{compute}
We conducted all experiments using a single NVIDIA H100 GPU. Reproducing the main results requires 65 GB of VRAM, while the NLP experiments used up to 75 GB.
\section{Main results with 1-sigma error bars}\label{sigmaapp}
In this section, we present extended versions of our main results with additional statistical details. All experiments were repeated using 5 different random seeds to account for variability. Tables~\ref{tab:sigma-one-mnist} and~\ref{tab:sigma-one-cifar} provide expanded versions of Table~\ref{tab:resultsOne}, reporting the mean and the corresponding 1-sigma error bars.

\begin{table}[H]
    \small
    \centering
   
    \begin{tabular}{c| c c c c c }
        \toprule
        \textbf{Method} & [\textbf{0,1}] & [\textbf{2,3}] & [\textbf{4,5}] & [\textbf{6,7}] &
        [\textbf{8,9}]\\
        \midrule
        \multirow{1}{*}{Regular FL}    & $\mathbf{96.18} {\scriptstyle \pm 0.85}$ & $\mathbf{79.25} {\scriptstyle \pm 5.91}$ & $\mathbf{66.84} {\scriptstyle \pm 8.34}$ & $\mathbf{88.12} {\scriptstyle \pm 3.65}$  & $\mathbf{66.38} {\scriptstyle \pm 12.18}$  \\
                               
        \midrule
      \multirow{1}{*}    {Dictator client: $1$}    & $\mathbf{99.63} {\scriptstyle \pm 0.11}$ & $\mathbf{0.00} {\scriptstyle \pm 00.00}$ & $\mathbf{0.00} {\scriptstyle \pm 00.00}$ & $\mathbf{0.00} {\scriptstyle \pm 00.00}$ & $\mathbf{0.00} {\scriptstyle \pm 00.00}$ \\
       \multirow{1}{*}    {Dictator client: $2$}    & $\mathbf{0.00} {\scriptstyle \pm 00.00}$ & $\mathbf{93.92} {\scriptstyle \pm 1.64}$ & $\mathbf{0.00} {\scriptstyle \pm 00.00}$ & $\mathbf{0.00} {\scriptstyle \pm 00.00}$ & $\mathbf{0.00} {\scriptstyle \pm 00.00}$  \\
       \multirow{1}{*}    {Dictator client: $3$}    & $\mathbf{0.00} {\scriptstyle \pm 00.00}$ & $\mathbf{0.00} {\scriptstyle \pm 00.00}$ & $\mathbf{97.43} {\scriptstyle \pm 0.99}$ & $\mathbf{0.00} {\scriptstyle \pm 00.00}$ & $\mathbf{0.00} {\scriptstyle \pm 00.00}$\\
       \multirow{1}{*}    {Dictator client: $4$}    & $\mathbf{0.00} {\scriptstyle \pm 00.00}$ & $\mathbf{0.00} {\scriptstyle \pm 00.00}$ & $\mathbf{0.00} {\scriptstyle \pm 00.00}$ & $\mathbf{98.91} {\scriptstyle \pm 0.68}$ & $\mathbf{0.00} {\scriptstyle \pm 00.00}$ \\
       \multirow{1}{*}    {Dictator client: $5$}    & $\mathbf{0.00} {\scriptstyle \pm 00.00}$ & $\mathbf{0.00} {\scriptstyle \pm 00.00}$ & $\mathbf{0.00} {\scriptstyle \pm 00.00}$ & $\mathbf{0.00} {\scriptstyle \pm 00.00}$ & $\mathbf{94.42} {\scriptstyle \pm 0.48}$  \\
       \midrule
      \multirow{1}{*}    {Dictator clients: $2$,$3$}    & $\mathbf{0.00} {\scriptstyle \pm 0.00}$ & $\mathbf{88.19} {\scriptstyle \pm 4.15}$ & $\mathbf{87.80} {\scriptstyle \pm 4.18}$ & $\mathbf{0.00} {\scriptstyle \pm 0.00}$  & $\mathbf{0.00} {\scriptstyle \pm 0.00}$   \\
       \multirow{1}{*}    {Dictator clients: $2$,$3$,$4$}   & $\mathbf{0.00} {\scriptstyle \pm 0.00}$ & $\mathbf{84.87} {\scriptstyle \pm 2.98}$ & $\mathbf{80.22} {\scriptstyle \pm 6.43}$ & $\mathbf{94.19} {\scriptstyle \pm 2.13}$  & $\mathbf{0.00} {\scriptstyle \pm 0.00}$  \\
    
        \bottomrule
    \end{tabular}
     \caption{Performance of the global model on each local dataset for MNIST and the single dictator client and collaborative dictator clients scenarios.}
    \label{tab:sigma-one-mnist}
\end{table}

\begin{table}[H]
    \small
    \centering
   
    \begin{tabular}{c| c c c c c }
        \toprule
        \textbf{Method} & [\textbf{0,1}] & [\textbf{2,3}] & [\textbf{4,5}] & [\textbf{6,7}] &
        [\textbf{8,9}]\\
        \midrule
        \multirow{1}{*}{Regular FL}    & $\mathbf{39.04} {\scriptstyle \pm 3.85}$ & $\mathbf{12.51} {\scriptstyle \pm 5.82}$ & $\mathbf{31.07} {\scriptstyle \pm 2.30}$ & $\mathbf{23.74} {\scriptstyle \pm 4.53}$  & $\mathbf{52.59} {\scriptstyle \pm 1.59}$  \\
                               
        \midrule
      \multirow{1}{*}    {Dictator client: $1$}    & $\mathbf{73.65} {\scriptstyle \pm 11.99}$ & $\mathbf{0.00} {\scriptstyle \pm 00.00}$ & $\mathbf{0.00} {\scriptstyle \pm 00.00}$ & $\mathbf{0.00} {\scriptstyle \pm 00.00}$ & $\mathbf{0.00} {\scriptstyle \pm 00.00}$ \\
       \multirow{1}{*}    {Dictator client: $2$}    & $\mathbf{0.00} {\scriptstyle \pm 00.00}$ & $\mathbf{65.19} {\scriptstyle \pm 8.65}$ & $\mathbf{0.00} {\scriptstyle \pm 00.00}$ & $\mathbf{0.00} {\scriptstyle \pm 00.00}$ & $\mathbf{0.00} {\scriptstyle \pm 00.00}$  \\
       \multirow{1}{*}    {Dictator client: $3$}    & $\mathbf{0.00} {\scriptstyle \pm 00.00}$ & $\mathbf{0.00} {\scriptstyle \pm 00.00}$ & $\mathbf{66.51} {\scriptstyle \pm 11.90}$ & $\mathbf{0.00} {\scriptstyle \pm 00.00}$ & $\mathbf{0.00} {\scriptstyle \pm 00.00}$\\
       \multirow{1}{*}    {Dictator client: $4$}    & $\mathbf{0.00} {\scriptstyle \pm 00.00}$ & $\mathbf{0.00} {\scriptstyle \pm 00.00}$ & $\mathbf{0.00} {\scriptstyle \pm 00.00}$ & $\mathbf{73.98} {\scriptstyle \pm 4.66}$ & $\mathbf{0.00} {\scriptstyle \pm 00.00}$ \\
       \multirow{1}{*}    {Dictator client: $5$}    & $\mathbf{0.00} {\scriptstyle \pm 00.00}$ & $\mathbf{0.00} {\scriptstyle \pm 00.00}$ & $\mathbf{0.00} {\scriptstyle \pm 00.00}$ & $\mathbf{0.00} {\scriptstyle \pm 00.00}$ & $\mathbf{77.06} {\scriptstyle \pm 4.79}$  \\
       \midrule
      \multirow{1}{*}    {Dictator clients: $2$,$3$}    & $\mathbf{0.00} {\scriptstyle \pm 0.00}$ & $\mathbf{35.08} {\scriptstyle \pm 18.92}$ & $\mathbf{43.17} {\scriptstyle \pm 17.73}$ & $\mathbf{0.00} {\scriptstyle \pm 0.00}$  & $\mathbf{0.00} {\scriptstyle \pm 0.00}$   \\
       \multirow{1}{*}    {Dictator clients: $2$,$3$,$4$}   & $\mathbf{0.00} {\scriptstyle \pm 0.00}$ & $\mathbf{18.38} {\scriptstyle \pm 10.77}$ & $\mathbf{40.02} {\scriptstyle \pm 8.37}$ & $\mathbf{46.05} {\scriptstyle \pm 5.85}$  & $\mathbf{0.00} {\scriptstyle \pm 0.00}$  \\
    
        \bottomrule
    \end{tabular}
     \caption{Performance of the global model on each local dataset for CIFAR10 and the single dictator client and collaborative dictator clients scenarios.}
    \label{tab:sigma-one-cifar}
\end{table}

\end{document}